\documentclass[twoside,twocolumn,journal]{IEEEtran}
\usepackage{amsmath,amsfonts}
\usepackage{algorithmic}
\usepackage{algorithm}
\usepackage{array}
\usepackage{textcomp}
\usepackage{stfloats}
\usepackage{url}
\usepackage{verbatim}
\usepackage{graphicx}
\usepackage{cite}
\usepackage{multirow}
\usepackage{pifont}
\usepackage{float}
\usepackage{color}
\usepackage{subfigure}
\usepackage{longtable}

\usepackage{hyperref}
\hypersetup{colorlinks=true}

\begin{document}

\title{
MTP: Advancing Remote Sensing Foundation Model via Multi-Task Pretraining
}

\author{
        Di Wang,~\IEEEmembership{Member,~IEEE,}
        Jing Zhang,~\IEEEmembership{Senior Member,~IEEE,}
        Minqiang Xu,
        Lin Liu, 
        Dongsheng Wang,\\
        Erzhong Gao,
        Chengxi Han,~\IEEEmembership{Student Member,~IEEE,}
        Haonan Guo,~\IEEEmembership{Student Member,~IEEE,}
        Bo Du,~\IEEEmembership{Senior Member,~IEEE,}
        Dacheng Tao,~\IEEEmembership{Fellow,~IEEE}
        and Liangpei Zhang,~\IEEEmembership{Fellow,~IEEE}

\thanks{D. Wang and B. Du are with the School of Computer Science, Wuhan University, Wuhan 430072, China, also with the Institute of Artificial Intelligence, Wuhan University, Wuhan 430072, China, also with the National Engineering Research Center for Multimedia Software, Wuhan University, Wuhan 430072, China, and also with the Hubei Key Laboratory of Multimedia and Network Communication Engineering, Wuhan University, Wuhan 430072, China (e-mail: wd74108520@gmail.com; dubo@whu.edu.cn).\textit{(Corresponding author: Minqiang Xu, Jing Zhang, Bo Du and Liangpei Zhang.)}}
\thanks{J. Zhang is with the School of Computer Science, Faculty of Engineering, The University of Sydney, Australia (e-mail: jing.zhang1@sydney.edu.au).}
\thanks{M. Xu, L. Liu, D. Wang and E. Gao are with the iFlytek Co., Ltd and also with the National Engineering Research Center of Speech and Language Information Processing, Hefei 230088, China (e-mail: mqxu7@iflytek.com; linliu@iflytek.com; dswang7@iflytek.com; ezgao@iflytek.com).}
\thanks{C. Han, H. Guo and  L. Zhang are with the State Key Laboratory of Information Engineering in Surveying, Mapping and Remote Sensing, Wuhan University, Wuhan 430079, China (e-mail: chengxihan@whu.edu.cn; haonan.guo@whu.edu.cn; zlp62@whu.edu.cn).}%
\thanks{D. Tao is with the School of Computer Science and Engineering, Nanyang Technological University, Singapore (e-mail: dacheng.tao@gmail.com).}
}

\markboth{Journal of \LaTeX\ Class Files,~Vol.~14, No.~8, August~2021}%
{Wang \MakeLowercase{\textit{et al.}}: ADVANCING RS Foundation model via MTP}


\maketitle

\begin{abstract}

Foundation models have reshaped the landscape of Remote Sensing (RS) by enhancing various image interpretation tasks. Pretraining is an active research topic, encompassing supervised and self-supervised learning methods to initialize model weights effectively. However, transferring the pretrained models to downstream tasks may encounter task discrepancy due to their formulation of pretraining as image classification or object discrimination tasks. In this study, we explore the Multi-Task Pretraining (MTP) paradigm for RS foundation models to address this issue. Using a shared encoder and task-specific decoder architecture, we conduct multi-task supervised pretraining on the SAMRS dataset, encompassing semantic segmentation, instance segmentation, and rotated object detection. MTP supports both convolutional neural networks and vision transformer foundation models with over 300 million parameters. The pretrained models are finetuned on various RS downstream tasks, such as scene classification, horizontal and rotated object detection, semantic segmentation, and change detection. Extensive experiments across 14 datasets demonstrate the superiority of our models over existing ones of similar size and their competitive performance compared to larger state-of-the-art models, thus validating the effectiveness of MTP. The codes and pretrained models will be released at https://github.com/ViTAE-Transformer/MTP.
\end{abstract}


\section{Introduction}

\IEEEPARstart{R}emote sensing (RS) image is one of the most important data resources for recording ground surfaces and land objects. Precisely understanding RS images is beneficial to many applications, including urban planning \cite{urban_plan}, environmental survey \cite{environ_survey}, disaster assessment \cite{disa_ass}, etc.

Utilizing its inherent capability to automatically learn and extract deep features from objects, deep learning methods have found widespread application in the RS domain, particularly for tasks such as scene classification, land use and land cover classification, and ship detection. Typically, ImageNet pretrained weights are employed in training deep networks for RS tasks due to their extensive representational ability. However, these weights are derived from pretraining models on natural images, leading to domain gaps between natural images and RS images. For instance, RS images are captured from a bird's-eye view, lack the vibrant colors of natural images, and possess lower spatial resolution. These disparities may impede the model's finetuning performance \cite{saumoco,rsp}. Moreover, relying solely on limited task-specific data for training restricts the model size and generalization capability of current RS deep models due to the notorious overfitting issue.

To tackle these challenges, the development of RS vision foundation models is imperative, which should excel in extracting representative RS features. However, the RS domain has long grappled with a scarcity of adequately large annotated datasets, impeding related investigations. Until recently, the most expansive RS scene labeling datasets were fMoW \cite{fmow} and BigEarthNet \cite{bigearthnet}, boasting 132,716 and 590,326 unique scene instances \cite{millionaid}, respectively — yet still falling short of benchmarks set by natural image datasets like ImageNet-1K \cite{imagenet}. Long et al. \cite{millionaid} addressed this gap by introducing MillionAID, a large-scale RS scene labeling dataset with a closed sample capacity of 100,0848 compared to ImageNet-1K, igniting interest in supervised RS pretraining \cite{asp, rsp}. These studies show the feasibility of pretraining RS foundation models on large-scale RS datasets. Nonetheless, supervised pretraining of RS foundation models may not be the most preferable choice due to the expertise and substantial time and labor costs associated with labeling RS images.

\begin{figure}[t]
  \centering
  \includegraphics[width=\linewidth]{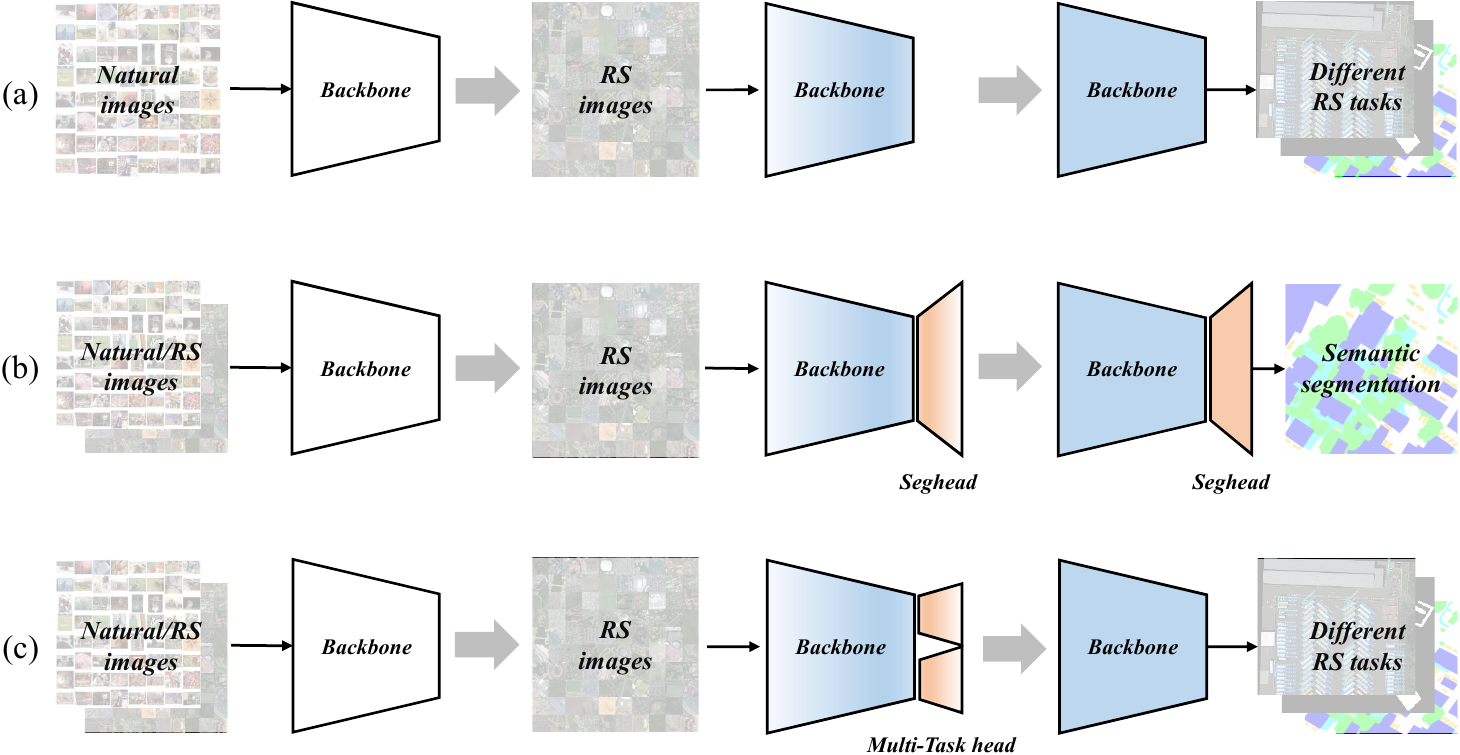}\\
  \caption{Comparison of various pretraining methods. (a) \cite{cspt, tov} sequentially pretrains a foundational model on both natural and RS images. (b) \cite{samrs} employs a two-stage pretraining strategy to initialize task-specific decoders (\textit{e.g.}, segmentation) using existing foundational models pretrained on either natural or RS images, preserving the decoder during subsequent finetuning. We extend (b) by incorporating multi-task decoders to enhance the representation capacity of the foundational model, facilitating easy transferability across diverse tasks during finetuning, as depicted in (c).
  }
  \label{compare_pipeline}
\end{figure}

Constructing large-scale RS annotation datasets is challenging due to the high complexity and cost of labeling. Despite this challenge, the advancement of earth observation technologies grants easy access to a vast amount of unlabeled RS images. Efficiently leveraging these unlabeled RS images is crucial for developing robust RS foundation models. In the realm of deep learning, unsupervised pretraining has emerged as a promising approach for learning effective knowledge from massive unlabeled data \cite{bert, gpt3, dino, clip}. Typically, unsupervised pretraining employs self-supervised learning (SSL) to learn effective feature representation. SSL encompasses two primary techniques: contrastive-based \cite{simclr, moco, byol} and generative-based learning \cite{beit, simmim, mae}. Contrastive learning aims to bring similar samples closer while maximizing distances between dissimilar samples through the object discrimination pretext task. When applied to the RS domain, data characteristics like geographic coordinates \cite{matter, csp, geoclip} and temporal information \cite{gassl, caco, seco} are usually leveraged in formulating the pretext task. However, designing these pretext tasks and gathering requisite data can be inefficient, especially for training large-scale models. Generative-based learning, exemplified by masked image modeling (MIM), circumvents this challenge by enhancing network representation through reconstructing masked regions. Many RS studies leverage MIM initialization for its efficiency \cite{rvsa, satmae, ringmo, ringmo_sense, spectralgpt, bfm, scale_mae, ctxmim}. Recent approaches have attempted to combine contrastive-based and generative-based learning techniques to pretrain more powerful models \cite{cmid, cross_mae, croma}. 

However, existing research usually resorts to a single data source. For instance, \cite{rsp, rvsa} utilize RGB aerial images from MillionAID, while \cite{satmae, spectralgpt} utilize Sentinel-2 multispectral images. Despite recent advancements in RS multimodal foundation models \cite{fgmae, skysense, decur, cmrsfm_mdrcdf}, which are beginning to incorporate more diverse imagery such as SAR, they still remain within the realm of in-domain data, namely pretraining with RS data. However, restricting pretraining solely to RS images may limit model capabilities since understanding RS objects requires specialized knowledge \cite{tov}. Can RS foundation models benefit from incorporating information from other data sources? \cite{rsp} suggests that traditional ImageNet pretraining aids in learning universal representations, whereas pretraining on RS data is particularly beneficial for recognizing RS-related categories. To address this, \cite{gersp} develop a teacher-student framework that integrates ImageNet supervised pretraining and RS unsupervised pretraining simultaneously, while \cite{gfm} employs representations from ImageNet to enhance the learning process of MIM for improving RS foundation models. Additionally, \cite{tov} and \cite{cspt} sequentially pretrain models on natural images and RS images using contrastive SSL or MAE \cite{mae}, respectively, as illustrated in Figure~\ref{compare_pipeline}(a).

While previous RS foundation models have shown remarkable performance across various RS tasks, a persistent challenge remains: the \textit{task discrepancy} between pretraining and finetuning, which often dictates the effectiveness of migrating pretrained models to downstream tasks. Research has highlighted the impact of representation granularity mismatch between pretraining and finetuning tasks \cite{rsp}. For instance, models pretrained on scene-level classification tasks perform favorably when finetuned on similar tasks but falter on pixel-level segmentation tasks. To address this issue, recent work \cite{samrs} has explored the segmentation pretraining paradigm, as shown in Figure~\ref{compare_pipeline}(b), yielding improved finetuning results. This suggests that enhancing model representation capability through additional pretraining, particularly on tasks demanding finer representation granularity, such as pixel-level segmentation, could be beneficial. Motivated by these findings, we ask: \textit{can we significantly enhance RS foundation models' representation ability through additional pretraining incorporating multiple tasks with diverse representation granularity?} To this end, we investigate the Multi-Task Pretraining (MTP) paradigm to bridge the gap between upstream and downstream tasks and obtain more powerful RS foundation models, as shown in Figure~\ref{compare_pipeline}(c). Importantly, MTP is designed to be applied to any existing pretraining models, irrespective of whether trained on RS or natural images.

Implementing MTP to bridge upstream-downstream task discrepancy necessitates the utilization of a similar or the same pretraining task as the downstream one, such as segmentation pretraining (SEP) for RS segmentation tasks \cite{samrs}. Therefore, to cover the common task types in typical downstream applications, MTP tasks should encompass dense prediction tasks like object detection and semantic segmentation. Hence, MTP requires a pretraining dataset with labels for these tasks, ideally, each sample encompassing all task labels. However, existing RS datasets often lack annotations for segmentation and rotated object detection. Fortunately, recent work \cite{samrs} introduces SAMRS, a large-scale segmentation dataset derived from existing RS rotated object detection datasets via the Segment Anything Model (SAM) \cite{sam}. SAMRS provides both detection and segmentation labels, facilitating MTP across RS semantic segmentation, instance segmentation, and rotated object detection tasks. Utilizing SAMRS, we demonstrate MTP's efficacy in enhancing RS foundation models, including both convolutional neural networks (CNN) and vision transformer foundation models with over 300 million parameters.

The main contributions of this paper are three-fold:
\begin{enumerate}
    \item [1)] We address the discrepancy between upstream pretraining and downstream finetuning tasks by introducing a stage-wise multi-task pretraining approach to enhance the RS foundation model.
    \item [2)] We utilize MTP to pretrain representative CNN and vision transformer foundation models with over 300M parameters on the SAMRS dataset, encompassing semantic segmentation, instance segmentation, and rotated object detection tasks in a unified framework.
    \item [3)] Extensive experiments demonstrate that MTP significantly advances the representation capability of RS foundation models, delivering remarkable performance across various RS downstream tasks such as scene classification, semantic segmentation, object detection, and change detection.
\end{enumerate}

The remainder of this paper is organized as follows. Section~\ref{sec:relatedwork} introduces the existing works related to supervised, multi-stage, and multi-task RS pretraining. Section~\ref{sec:mtp} presents the details of MTP, where the used SAMRS dataset and vision foundation models are also briefly introduced. Experimental results and corresponding analyses are depicted in Section~\ref{sec:experiments}. Finally, Section~\ref{sec:conclusion} concludes this paper.

\section{Related Work}
\label{sec:relatedwork}
\subsection{Supervised Pretraining for RS Foundation Model}

Before the rise of SSL-based RS foundation models, researchers have already delved into pretraining deep models using labeled RS datasets. Tong et al. \cite{gid} pretrained an ImageNet-pretrained ResNet-50 \cite{resnet} using images from the GID dataset \cite{gid} to derive pseudo-labels for precise land-cover classification on high-resolution RS images. Recognizing the challenge of labeling large-scale RS images, others sought alternatives to RS annotation datasets. For instance, Li et al. \cite{geokr} utilized the global land cover product Globeland30 \cite{globeland30} as supervision for RS representation learning. They adopted a mean-teacher framework to mitigate random noise stemming from inconsistencies in imaging time and resolution between RS images and geographical products. Moreover, they incorporated additional geographical supervisions, such as change degree and spatial aggregation, to regularize the pretraining process \cite{geco}. Long et al. \cite{asp} subsequently demonstrated the effectiveness of various CNN models (including AlexNet \cite{alexnet}, VGG-16 \cite{vgg}, GoogleNet \cite{googlenet}, ResNet-101 \cite{resnet}, and DenseNet-121/169 \cite{densenet}) pretrained from scratch on the MillionAID dataset. Their models outperformed traditional ImageNet pretrained models in scene classification tasks, indicating the potential of leveraging large-scale RS datasets for pretraining. Later, Wang et al. \cite{rsp} pretrained typical CNN models and vision transformer models, including Swin-T \cite{swint} and ViTAEv2 \cite{vitae_v2}, all randomly initialized, on the MillionAID. They conducted a comprehensive empirical study comparing finetuning performance using different pretraining strategies (MillionAID vs. ImageNet) across four types of RS downstream tasks: scene recognition, semantic segmentation, rotated object detection, and change detection. Their results demonstrated the superiority of vision transformer models over CNNs on RS scenes and validated the feasibility of constructing RS foundation models via supervised pretraining on large-scale RS datasets. Bastani et al. \cite{satlas} introduced the larger Satlas dataset for RS supervised pretraining. Very recently, SAMRS \cite{samrs} introduced supervised semantic segmentation pretraining to enhance model performance on the segmentation task. Inspired by \cite{samrs}, this paper revisits the supervised learning approach by integrating it with existing pretraining strategies, such as ImageNet pretraining, and exploring multi-task pretraining to construct distinct RS foundation models.

\subsection{Multi-Stage Pretraining for RS Foundation Model}
 
Given the domain gap between RS images and natural images or between various RS modalities, it is reasonable to conduct multiple rounds of pretraining. Gururangan et al. \cite{dont_stop_trn} demonstrated that unsupervised pretraining on in-domain or task-specific data enhances model performance in natural language processing (NLP) tasks. Building on this insight, Zhang et al. \cite{cspt} devised a sequential pretraining approach, initially on ImageNet followed by the target RS dataset, employing MIM for pretraining. Similarly, \cite{tov} proposed a strategy inspired by human-like learning, first performing contrastive SSL on natural images, then freezing shallow layer weights and conducting SSL on an RS dataset. Contrary to \cite{dont_stop_trn}, Dery et al. \cite{end_task_aware_trn} introduced stronger end-task-aware training for NLP tasks by integrating auxiliary data and end-task objectives into the learning process. Similarly, \cite{samrs} introduced additional segmentation pretraining using common segmenters (\textit{e.g.}, UperNet \cite{upernet} and Mask2Former \cite{mask2former}) and the SAMRS dataset, enhancing model accuracy in RS segmentation tasks. Notably, our objective diverges from \cite{samrs} in applying stage-wise pretraining. While \cite{samrs} retains the segmentor after segmentation pretraining to enhance segmentation performance, we aim to enhance the representation capability of RS foundation models via stage-wise pretraining, preserving only the backbone network after pretraining to facilitate transfer to diverse RS downstream tasks.

\subsection{Multi-Task Pretraining for RS Foundation Model}

Applying multi-task learning to enhance the RS foundation model is an intuitive idea. Li et al. \cite{ssmtl_semseg} introduced multi-task SSL representation learning, combining image inpainting, transform prediction, and contrast learning to boost semantic segmentation performance in RS images. However, it was limited to finetuning a pretrained model solely on semantic segmentation tasks, constrained by model size and pretraining dataset capacity. RSCoTr \cite{rscotr} constructs a multi-task learning framework to simultaneously achieve classification, segmentation, and detection tasks. Unfortunately, the network can only be optimized by one task in each iteration during training due to lacking of multi-label datasets. The aspiration to consolidate multiple tasks into a single model has been a longstanding pursuit \cite{gpt3, clip, florence, nuwa, blip, glip, coca, beit3, xu2021vitae, vitae_v2, qformer, rsgpt, skysense, fcd_gan}, aligning with the original goals of the foundation model exploration. Bastani et al. \cite{satlas} devised a multi-task model by integrating Swin-Base \cite{swint} with seven heads from existing networks (\textit{e.g.}, Faster-RCNN \cite{FasterRCNN} and UNet \cite{unet}), facilitating training on the multi-task annotated Satlas dataset. However, their approach lacked incorporation of typical RS rotated object tasks, focusing solely on transferring the model to RS classification datasets. Inspired by these pioneering efforts, this paper constructs a unified multi-task pretraining framework that supports multiple datasets, where the sample in each dataset simultaneously possesses multi-task labels, which are uniformly processed in a data loader pipeline. During the training, the model can be simultaneously optimized through multiple tasks when in each iteration. Based on this framework, we pretrain RS foundation models with over 300M parameters, encompassing semantic segmentation, instance segmentation, and rotated object detection tasks using the SAMRS dataset. After pretraining, the backbone network is further finetuned on various RS downstream tasks.

\section{Multi-Task Pretraining}
\label{sec:mtp}

We utilize semantic segmentation, instance segmentation, and rotated object detection annotations from the SAMRS dataset for Multi-Task Pretraining (MTP). Advanced CNN and vision transformer models serve as the backbone networks to thoroughly investigate MTP. This section begins with an overview of the SAMRS dataset, followed by a brief introduction to the selected models. Subsequently, we present the MTP framework and implementation details.

\subsection{SAMRS Dataset}

SAMRS (\textbf{S}egment \textbf{A}nything \textbf{M}odel annotated \textbf{R}emote Sensing \textbf{S}egmentation dataset) \cite{samrs} is a large-scale RS segmentation database, comprising 105,090 images and 1,668,241 instances from three datasets: SOTA, SIOR, and FAST. These datasets are derived from existing large-scale RS object detection datasets, namely DOTA-V2.0 \cite{dotav2}, DIOR \cite{dior}, and FAIR1M-2.0 \cite{fair1m}, through transforming the bounding box annotations using the SAM \cite{sam}. SAMRS inherits the categories directly from the original detection datasets, resulting in a capacity exceeding that of most existing RS segmentation datasets by more than tenfold (\textit{e.g.}, ISPRS Potsdam\footnote{https://www.isprs.org/education/benchmarks/UrbanSemLab/2d-sem-labelpotsdam.aspx} and LoveDA \cite{loveda}). The image sizes for the three sets are 1,024 $\times$ 1,024, 800 $\times$ 800, and 600 $\times$ 600, respectively. Despite being primarily intended for large-scale pretraining exploration rather than benchmarking due to its automatically generated labels, SAMRS naturally supports instance segmentation and object detection. This versatility extends its utility to investigating large-scale multi-task pretraining.

\begin{figure*}[ht]
  \centering
  \includegraphics[width=\linewidth]{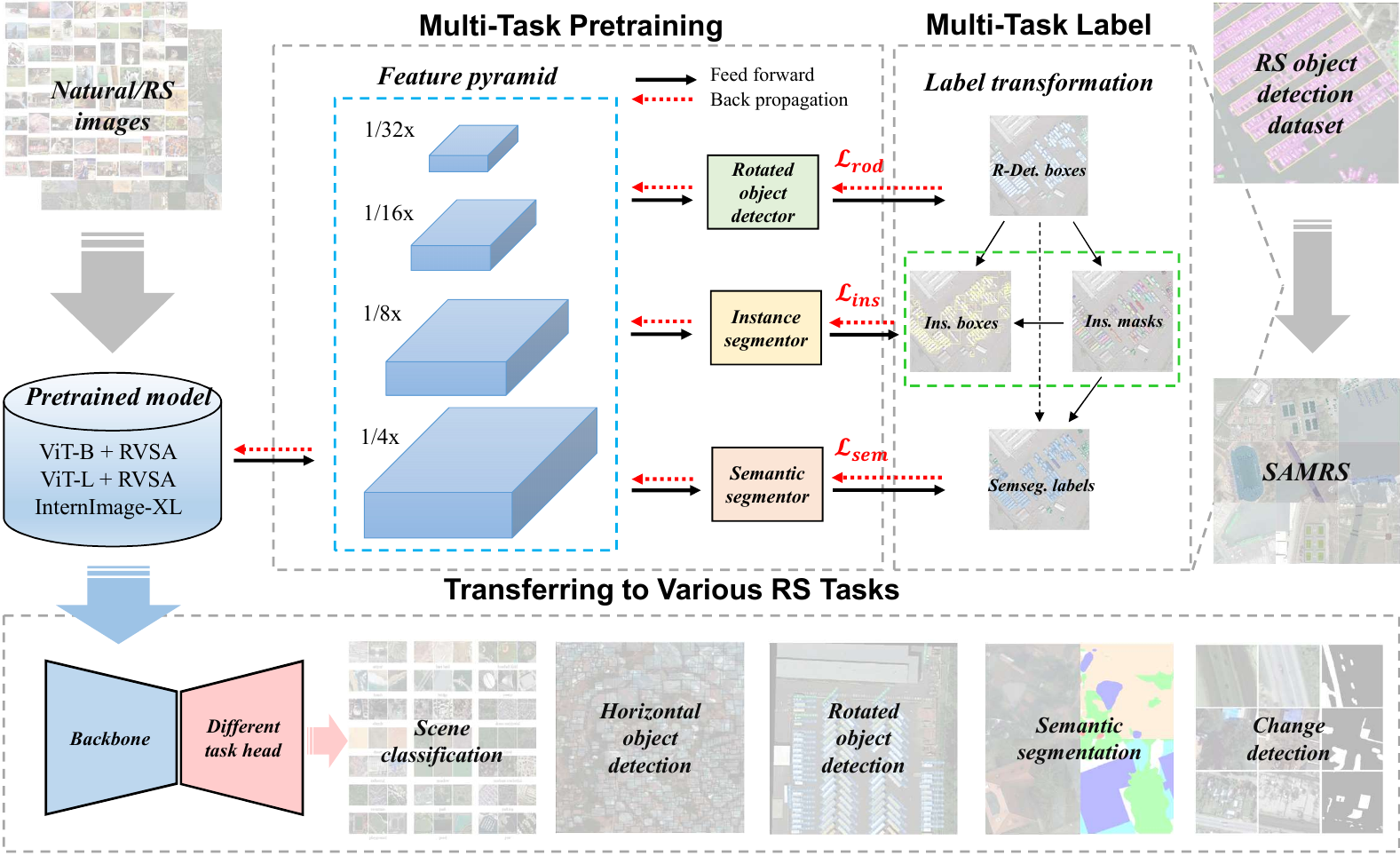}\\
  \caption{
  The overall pipeline of MTP. Inside MTP, the feature pyramid from the backbone network is fed into multiple decoders for various tasks, including rotated object detection, instance segmentation, and semantic segmentation. These tasks are supervised by diverse labels in the SAMRS dataset. Following MTP, the pretrained model is transferred to different RS tasks for finetuning.
  }
  \label{mtp_framework}
\end{figure*}

\subsection{Backbone Network}
In this research, we adopt RVSA (\textbf{R}otated \textbf{V}aried-\textbf{S}ize window \textbf{A}ttention) \cite{rvsa} and InternImage \cite{internimage} as the representative vision transformer-based and CNN-based foundation models.

\subsubsection{RVSA} This model is specially designed for RS images. Considering the various orientations of RS objects caused by the bird's-eye view, this model extends the varied-size window attention in \cite{vsa} by additionally introducing a learnable angle factor, offering windows that can adaptively zoom, translate, and rotate.

Specifically, given an input feature $\mathbf{X} \in\mathbb{R}^{C\times H\times W}$ ($C, H, W$ are the number of channel, height, and width in $\mathbf{X}$), which is evenly divided into different windows, where the feature of each window can be formulated as $\mathbf{X}_w \in \mathbb{R}^{C\times s \times s}$ ($s$ is the window size), obtaining $\frac{H}{s}\times\frac{W}{s}$ windows totally. Then, three linear layers are used to generate the query feature, and initial key and value features, which are separately represented as $\mathbf{Q}_w$, $\mathbf{K}_w$ and $\mathbf{V}_w$. We use $\mathbf{X}_w$ to predict the variations of the window:
  \begin{equation}
  S_w, O_w, \Theta_w = Linear(LeakyReLU(GAP(\mathbf{X}_w)))
  \end{equation}
  Here, GAP is global average pooling, $S_w = \{s_x, s_y \in \mathbb{R}^1 \}$ and $O_w = \{o_x, o_y \in \mathbb{R}^1\}$ are the scale factor and offset in the X and Y axis, while $\Theta_w = \{\theta \in \mathbb{R}^1\}$ is the rotation
  angle. Taking an example using the corner points of a window: 
  \begin{equation}
  \left[\begin{array}{c}
    x_{l/r} \\
    y_{l/r} \\
  \end{array}\right]
  = \left[\begin{array}{c}
    x^c \\
    y^c \\
  \end{array}\right] + \left[\begin{array}{c}
    x_{l/r}^r \\
    y_{l/r}^r \\
  \end{array}\right]
\end{equation}
where $x_l, y_l, x_r, y_r$ are the coordinates of the upper left
and lower right corners of the initial window, $x_c, y_c$ are the coordinates of the window center point. Therefore, $x_l^r, y_l^r, x_r^r, y_r^r$ are the distances between the corner points and the center in horizontal and vertical directions. The transformation of the window can be implemented using the obtained scaling, translation, and rotation factors:
\begin{equation}
  \begin{split}
    \left[\begin{array}{c}
      x_{l/r}' \\
      y_{l/r}' \\
    \end{array}\right]
    & = \left[\begin{array}{c}
      x^c \\
      y^c \\
    \end{array}\right]
    + \left[\begin{array}{c}
      o_x \\
      o_y \\
    \end{array}\right] \\
    & + \left[\begin{array}{cc}
      \cos{\theta} & \sin{\theta} \\
      -\sin{\theta} & \cos{\theta} \\
    \end{array}\right] 
    \left[\begin{array}{c}
      x_{l/r}^r \cdot s_x\\
      y_{l/r}^r \cdot s_y\\
    \end{array}\right]
  \end{split}
\end{equation}
$x_l', y_l', x_r', y_r'$ are the coordinates of the corner points of the transformed window. Then, new key and value feature $\mathbf{K}_{w'}$ and $\mathbf{V}_{w'}$ can be sampled from the obtained window, and the self-attention (SA) can be operated by the following formula:
\begin{equation}
  \mathbf{F}_{w}= SA(\mathbf{Q}_w,\mathbf{K}_{w'},\mathbf{V}_{w'}) 
  =softmax(\frac{\mathbf{Q}_w \mathbf{K}_{w'}^T}{\sqrt{C'}}) \mathbf{V}_{w'}
\end{equation}
In this formula, $\mathbf{F}_{w} \in \mathbb{R}^{s^2 \times C'}$ is the feature output by one self-attention of one window, $C = hC'$, where $h$ is the number of SA. The shape of final output features in RVSA can be recovered by concatenating the features from different SAs in the channel dimension and merging features from different windows along the spatial dimension.  

RVSA is used to replace the original multi-head full attention in original vision transformers. To achieve a trade-off between accuracy and efficiency, following \cite{vitdet}, only the full attention in  1/4 depth layer is preserved. In the original paper \cite{rvsa}, RVSA is separately used on ViT \cite{vit} and ViTAE \cite{xu2021vitae}, whereas ViTAE is a CNN-Transformer hybrid model. In this paper, we employ the ViT-based version to investigate the impact of MTP on a plain vision transformer. In addition, the RVSA model in the original paper is limited to the base version of vision transformers, \textit{i.e.}, ViT-B + RVSA. To pretrain larger models, we further apply RVSA to ViT-Large, obtaining ViT-L + RVSA. Their detailed configurations are presented in Table~\ref{rvsa}.

\subsubsection{InternImage}

This model integrates the strengths of recent vision transformers and large kernels into CNNs via dynamic sparse kernels, combining long-range context capture, adaptive spatial information aggregation, and efficient computation. It extends deformable convolution \cite{dcn, dcnv2} with depth-wise and multi-head mechanisms and incorporates modern transformer designs such as layer normalization \cite{layernorm}, feed-forward networks \cite{selfattention}, and GELU activation \cite{gelu}. We evaluate its performance on diverse RS downstream tasks, showcasing its potential beyond its initial design for natural images. Furthermore, this choice facilitates investigating the impact of MTP on CNN-based models. Here, we employ the XL version to match the model size of ViT-L + RVSA.

\begin{table}[t]
  \caption{Detailed configurations of different RVSA models.}
  \centering
  \begin{tabular}{l|cc}
  \hline
 \bfseries Backbone & \bfseries ViT-B + RVSA & \bfseries ViT-L + RVSA \\
  \hline
  Depth & 12 & 24 \\
  Embedding Dim & 768 & 1024 \\
  Head & 12 &  16 \\
  Full attention Index & [3, 6, 9, 12] & [6, 12, 18, 24] \\
  Feature Pyramid Index & [4, 6, 8, 12] & [8, 12, 16, 24] \\
  \hline
\end{tabular}
 \label{rvsa}
\end{table}

\subsection{Multi-Task Pretraining}

We examine MTP using three models: ViT-B + RVSA, ViT-L + RVSA, and InternImage-XL. As the original RVSA research \cite{rvsa} focuses solely on the base version, we independently pretrain ViT-L on MillionAID similar to ViT-B + RVSA. These pretrained weights will be publicly accessible. Figure~\ref{mtp_framework} shows the overall pipeline of MTP. Technically, we further train the pretrained model on the SAMRS dataset, encompassing various annotations such as semantic segmentation, instance segmentation, and rotated object detection tasks concurrently. We employ well-established classical networks, including UperNet \cite{upernet}, Mask-RCNN \cite{maskrcnn}, and Oriented-RCNN \cite{orcn}, as segmentors or detectors. These networks utilize feature pyramids and are supervised with different labels. To illustrate this process, we depict the label transformation when generating SAMRS. Initially, rotated detection boxes (R-Det. boxes) are transformed into binary masks using SAM, serving as instance-level mask annotations. Subsequently, the minimum circumscribed horizontal rectangle of the binary mask is derived as instance-level box annotations, with categories inherited from rotated boxes. These instance-level annotations are utilized for instance segmentation. Semantic segmentation labels are then obtained by assigning rotated box categories to the masks. The losses stemming from these labels are $\mathcal{L}_{rod}$, $\mathcal{L}_{ins}$, and $\mathcal{L}_{sem}$, employed for the respective tasks. Notably, the instance segmentation loss comprises two components: the box annotation loss $\mathcal{L}_{ins\_b}$ and the binary mask loss $\mathcal{L}_{ins\_m}$. The overall loss for MTP is: 
\begin{equation}
    \mathcal{L} = \mathcal{L}_{rod} + \mathcal{L}_{ins\_b} + \mathcal{L}_{ins\_m} + \mathcal{L}_{sem}.
    \label{loss}
\end{equation}
Since SAMRS contains three sets, we have
\begin{equation}
\mathcal{L} = \sum_{i=1}^{3} \mathcal{L}_{rod}^{i} + \mathcal{L}_{ins\_b}^{i} + \mathcal{L}_{ins\_m}^{i} + \mathcal{L}_{sem}^{i},
\end{equation}
where $i$ indexes the three sub-sets: SOTA, SIOR, and FAST. The other settings of the loss follow the original papers \cite{upernet, orcn, maskrcnn}. In practice, we implement the overall framework based on MMSegmentation\footnote{https://github.com/open-mmlab/mmsegmentation}, MMDetection\footnote{https://github.com/open-mmlab/mmdetection}, and MMRotate\footnote{https://github.com/open-mmlab/mmrotate}. However, all these packages only support a single task. So we integrate the key components from these packages, such as the dataloader, model structure, loss function, and metric calculator, into a unified pipeline, to realize the MTP.

\subsection{Implementation Details}

\begin{table}[t]
  \caption{The training costs of implementing MTP using different models.}
  \newcommand{\tabincell}[2]{\begin{tabular}{@{}#1@{}}#2\end{tabular}}
  \centering
  \begin{tabular}{l|ccc}
  \hline
  \bfseries Backbone  & \bfseries \#Param.(M) &\bfseries  \#GPU &\bfseries  Time (days) \\
  \hline
  ViT-B + RVSA  & 86 & 16 & 3.0 \\
  ViT-L + RVSA  & 305 & 32 & 6.3 \\
  InternImage-XL & 335 & 32 & 6.3 \\
  \hline
\end{tabular}
 \label{trn_cost}
\end{table}

The pretraining is conducted on NVIDIA V100 GPUs. All models are trained for 80K iterations using the AdamW optimizer \cite{adamw}. The base learning rates of RVSA and InternImage are 0.00006 and 0.00002, respectively, with a weight decay of 0.05. We adopt an iteration-wise cosine annealing scheduler to adjust the learning rate. The layer decay rates of RVSA models and InternImage are 0.9 and 0.94, following original papers \cite{rvsa, internimage}. For ViT-B + RVSA, the batch size and input image size are set to 48 and 224, which are doubled for training larger models. Table \ref{trn_cost} lists the training costs of implementing MTP using different models.

\section{Finetuning Experiments}
\label{sec:experiments}

In this section, we thoroughly evaluate MTP's performance by finetuning pretrained models across four classical RS tasks: scene classification, object detection, semantic segmentation, and change detection, where the most representative and widely used benchmarks in the literature for different downstream tasks are employed for comparison. We also investigate the characteristics of MTP-based RS foundation models, examining the relationships between adopted datasets, hyperparameters, and finetuning performances, measuring accuracy variations with reduced training samples, and visualizing the predicted results.

\subsection{Scene Classification}

We first evaluate the pretrained models on the scene classification task. It does not need any extra decoder and can reflect the overall representation capability of the pretrained model.

\subsubsection{Dataset} We adopt two classical datasets: EuroSAT \cite{eurosat} and RESISC-45 \cite{asr_review} for scene classification. 

\begin{enumerate}
    \item[1)] EuroSAT: This dataset is captured by Sentinel-2 from Europe for land use and land cover classification. It has 10 classes, a total of 27,000 images with a resolution of 64 $\times$ 64. We adopt the public train/val split \cite{eurosat_split} by following \cite{seco, satmae}. 
    \item[2)] RESISC-45: This is a commonly-used dataset. It contains 31,500 images in a size of 256 $\times$ 256 across 45 categories, where each category possesses 700 samples. Following \cite{rsp, ringmo, rvsa, gersp, skysense}, we randomly select 20\% of the data for training and 80\% of the data for testing.
\end{enumerate}

\subsubsection{Implementation Details} 
In the implementation, all models are trained with a batch size of 64. The training epochs for EuroSAT and RESISC-45 are set to 100 and 200, respectively. The AdamW optimizer is used, where the base learning rate for RVSA and InterImage are 0.00006 and 0.00002, respectively, with a weight decay of 0.05. In the first 5 epochs, we adopt a linear warming-up strategy, where the initial learning rate is set to 0.000001. Then, the learning rate is controlled by the cosine annealing scheduler. The layer decay rates are 0.9 and 0.94 for RVSA and InternImage models, respectively. For classification, a global pooling layer and a linear head are used after the backbone network. To avoid overfitting, we adopt multiple data augmentations, including random resized cropping, random flipping, RandAugment \cite{randaugment}, and random erasing. Since the original image size of EuroSAT is too small, before feeding into the network, we resize the image to 224 $\times$ 224. The overall accuracy (OA) is used as the evaluation criterion. All experiments are implemented by MMPretrain\footnote{https://github.com/open-mmlab/mmpretrain}.

\subsubsection{Ablation Study of Stage-wise Pretraining}

\begin{table}[t]
  \caption{The OA(\%) of different model pretraining strategies on EuroSAT.}
  \centering
  \begin{tabular}{l|cc|c}
  \hline
  \bfseries  Model & \bfseries  MAE & \bfseries MTP  &\bfseries  OA   \\
  \hline 
  ViT-B + RVSA & \ding{52} &   &98.54  \\
  ViT-B + RVSA &  & \ding{52} & 98.24 \\
  ViT-B + RVSA & \ding{52} &  \ding{52} & 98.76 \\
  \hline
\end{tabular}
 \label{pretrn_ablation}
\end{table}

\begin{table}[t]
  \caption{The OA (\%) of finetuning different pretrained models on EuroSAT and RESISC-45 datasets. Here IMP means pretraining on ImageNet-22K.\textbf{bold} indicates the better accuracy of pretrained models with or without MTP. * represents the first three methods among all comparison methods, where the first, second, and third places are emphasized by black, \textcolor{red}{red} and \textcolor{green}{green} colors, respectively. }
  \newcommand{\tabincell}[2]{\begin{tabular}{@{}#1@{}}#2\end{tabular}}
  \centering
  \begin{tabular}{llcc}
  \hline
  \bfseries  Method &\bfseries  Model & \bfseries EuroSAT & \bfseries RESISC-45 \\
  \hline
  GASSL  \cite{gassl} & ResNet-18 & 89.51 & - \\
  SeCo \cite{seco}  & ResNet-18 &93.14 & - \\
  SatMAE \cite{satmae} & ViT-L  &  95.74  & 94.10 \\
  SwiMDiff \cite{swimdiff} & ResNet-18 & 96.10 &  - \\
  GASSL  \cite{gassl} & ResNet-50 & 96.38 & 93.06 \\
  GeRSP \cite{gersp} & ResNet-50 & - & 92.74 \\
  SeCo \cite{seco} & ResNet-50 & 97.34 & 92.91 \\
  CACo \cite{caco} & ResNet-50 &  97.77 &  91.94 \\
  TOV \cite{tov} & ResNet-50 & - & 93.79 \\
  RSP \cite{rsp} & ViTAEv2-S & - & 95.60 \\
  RingMo \cite{ringmo} & Swin-B & - & 95.67 \\
  SatLas \cite{satlas} & Swin-B & - & 94.70 \\
  CMID \cite{cmid} & Swin-B & - & 95.53 \\
  GFM \cite{gfm} & Swin-B & - & 94.64 \\
  CSPT \cite{cspt} & ViT-L & - & 95.62 \\
  Usat \cite{usat} & ViT-L & 98.37 & \\
  Scale-MAE \cite{scale_mae}  & ViT-L & 98.59 & 95.04 \\
  CtxMIM \cite{ctxmim} & Swin-B & 98.69 & - \\
  SatMAE++ \cite{satmae_pp} &ViT-L & 99.04 & - \\
  SpectralGPT$^+$ \cite{spectralgpt} & ViT-B & 99.21\textcolor{green}{*} & - \\
  SkySense \cite{skysense} & Swin-L & - & 95.92\textcolor{green}{*} \\
  SkySense \cite{skysense} & Swin-H & - & 96.32* \\
  \hline
  MAE &  ViT-B + RVSA  & 98.54  & 95.49 \\
  MAE + MTP & ViT-B + RVSA & \bfseries 98.76  & \bfseries 95.57 \\
   \hline
  MAE  & ViT-L + RVSA & 98.56 &  95.46 \\
  MAE + MTP & ViT-L + RVSA  & \bfseries 98.78 & \bfseries 95.88 \\
  \hline
  IMP & InternImage-XL & \bfseries 99.30*  & 95.82 \\
  IMP + MTP &  InternImage-XL & 99.24\textcolor{red}{*}  & \bfseries 96.27\textcolor{red}{*} \\
  \hline
\end{tabular}
\label{scene_cls}
\end{table}

As aforementioned, MTP is implemented based on existing pretraining models since it tries to address the task-level discrepancy. So an interesting question naturally arises: what about conducting MTP from scratch? To this end, we experiment by exploring different pretraining strategies using ViT-B + RVSA on EuroSAT, and the results are shown in Table \ref{pretrn_ablation}. It can be seen that, without using pretrained weights, MTP cannot achieve the ideal performance and even performs worse than MAE pretraining. These results demonstrate the importance of performing stage-wise pretraining.

\subsubsection{Finetuning Results and Analyses}
Table \ref{scene_cls} shows the finetuning results. It can be seen that MTP can improve existing foundation models on scene classification tasks, especially for the RVSA series. It helps the model achieve state-of-the-art performances compared to other pretraining models that have comparable sizes. With the help of MTP, on the RESISC-45 dataset, InterImage-XL surpasses Swin-L-based SkySense \cite{skysense}, which is pretrained on a tremendously large dataset that has more than 20 million multimodal RS image triplets involving RGB high-resolution images and multi-temporal multispectral and SAR sequences. MTP boosts the performance of InterImage-XL close to the Swin-H-based SkySense (96.27 v.s. 96.32), which has more parameters. We also notice the accuracy of IMP-InterImage-XL is decreased marginally in EuroSAT after MTP. We will investigate this phenomenon later. Nevertheless, the obtained model still outperforms SpectralGPT$^+$, which is pretrained with 1 million multispectral images, where each sample can be regarded as containing multiple groups of tri-spectral images, similar to RGB channels.

\subsection{Horizontal Object Detection}

After completing the scene-level task of recognition, we focus on the object-level tasks, \textit{i.e.}, horizontal and rotated object detection. Here, we first consider the horizontal detection task. 

\subsubsection{Dataset} We use two public datasets Xview \cite{xview} and DIOR \cite{dior} for evaluation. Here are the details:
\begin{enumerate}
\item[1)] Xview: This dataset is from the DIUx xView 2018 Detection Challenge \cite{xview}. It collects Worldview-3 satellite imagery beyond 1,400 $km^2$ in a ground resolution of 0.3$m$, involving 60 classes over 1 million object instances. Due to only the 846 images (beyond 2,000 $\times$ 2,000 pixels) in the training set are available, following  \cite{gassl, ctxmim}, we randomly select 700 images as the training set and 146 images for testing.
\item[2)] DIOR: This dataset consists of 23,463 images with resolutions ranging from 0.5 to 30$m$, including 192,472 instances. The images have been clipped to 800 $\times$ 800 for the convenience of model training and testing. It involves 20 common object categories. The training set, validation set, and testing set contain 5862, 5863, and 11738 samples, respectively. In this paper, we jointly use the training set and the validation set to finetune models and conduct the evaluation on the testing set.
\end{enumerate}

\subsubsection{Implementation Details} For Xview, we train a RetinaNet \cite{retinanet} by following \cite{ctxmim, gassl} with the pretrained model for 12 epochs, with a batch size of 8. While Faster-RCNN \cite{FasterRCNN} is adopted when finetuning on DIOR with the same settings except for a batch size of 4. We also apply a linear warming-up strategy with an initial learning rate of 0.000001 at the beginning of 500 iterations. We keep the same layer decay rates as the scene classification task. The basic learning rate, optimizer, and scheduler are the same as \cite{rvsa}. Before input into the network, the large images are uniformly clipped to 416 $\times$ 416 pixels. The data augmentation only includes random flipping with a probability of 0.5. We use MMDetection to implement the finetuning, where the $AP_{50}$ is used as the evaluation metric for the comparison of different models.

\begin{table}[t]
  \caption{The $AP_{50}$ (\%) of finetuning different pretrained models with RetinaNet on Xview and DIOR datasets. The ``Sup. Lea. w IN1K'' means supervised learning with ImageNet-1K. Random Init. means the backbone network is randomly initialized.}
  \newcommand{\tabincell}[2]{\begin{tabular}{@{}#1@{}}#2\end{tabular}}
  \centering
  \begin{tabular}{lllcc}
  \hline
  \bfseries Method & \bfseries Backbone & \bfseries Xview & \bfseries  DIOR \\
  \hline
  Random Init. &  ResNet-50  & 10.8  & - \\
  Sup. Lea. w IN1K & ResNet-50  & 14.4 &  -\\
  Sup. Lea. w IN1K & Swin-B   & 16.3  & - \\
  GASSL \cite{gassl} & ResNet-50 & 17.7 & 67.40 \\
  SeCo \cite{seco} &  ResNet-50 & 17.2 & - \\
  CACO \cite{caco} &  ResNet-50  & 17.2 &  66.91 \\
  CtxMIM \cite{ctxmim} & Swin-B  & 18.8\textcolor{red}{*} & -  \\
  MSFCNet \cite{msfcnet} & ResNeSt-101 \cite{resnest} & - & 70.08  \\
  TOV \cite{tov} & ResNet-50 & - & 70.16 \\
  SATMAE \cite{satmae} & ViT-L & - & 70.89 \\
  CSPT\cite{cspt} & ViT-L & - & 71.70 \\
  FSoDNet \cite{fsodnet} & MSENet & - & 71.80 \\
  GeRSP \cite{gersp} & ResNet-50 & - & 72.20 \\
  GFM \cite{gfm} & Swin-B & - & 72.84 \\
  Scale-MAE \cite{scale_mae} & ViT-L & - & 73.81 \\
  SatLas \cite{satlas} & Swin-B & - & 74.10 \\
  CMID \cite{cmid} & Swin-B & - & 75.11 \\
  RingMo \cite{ringmo} & Swin-B & - & 75.90 \\
  SkySense \cite{skysense} & Swin-H & - & 78.73\textcolor{green}{*} \\
 \hline
  MAE & ViT-B + RVSA & 14.6  & 75.80 \\
  MAE + MTP &  ViT-B + RVSA & \bfseries 16.4  & \bfseries 79.40\textcolor{red}{*} \\
  \hline
  MAE & ViT-L + RVSA & 15.0 & 78.30 \\
  MAE + MTP &   ViT-L + RVSA  & \bfseries 19.4*  & \bfseries 81.10* \\
  \hline
  IMP & InternImage-XL & 17.0 & 77.10  \\
  IMP + MTP & InternImage-XL & \bfseries 18.2\textcolor{green}{*}  & \bfseries 78.00 \\
  \hline
\end{tabular}
\label{hori_det}
\end{table}

\subsubsection{Finetuning Results and Analyses}
The experimental results are shown in Table \ref{hori_det}. We can find that the MTP enhances the performance of all pretrained models, especially for ViT-L + RVSA. On Xview, the performance of MAE pretrained ViT-L + RVSA is not as good as InterImage-XL, even worse than the smaller ResNet-50-based models. After utilizing MTP, the performance of ViT-L + RVSA has been greatly improved. It outperforms CtxMIM \cite{ctxmim} and achieves the best. On DIOR, with the help of MTP, ViT-B + RVSA has outperformed all existing methods, including the recently distinguished method SkySense \cite{skysense} that employs a larger model. In addition, MTP also greatly enhances ViT-L + RVSA, setting a new state-of-the-art. \textit{Here, we emphasize that despite the pretraining dataset SAMRS includes the samples of DIOR \cite{dior}. To avoid unfair comparison, following \cite{samrs}, the images of the testing set in DIOR have not been used for MTP. This rule also applies to other datasets that form the SAMRS, involving DOTA-V1.0 \cite{dota1}, DOTA-V2.0 \cite{dotav2}, DIOR-R \cite{aod_2022_tgrs_dior_r_aopg} and FAIR1M-2.0 \cite{fair1m}.} It should also be noted that the RVSA model is initially proposed by considering the diverse orientations of RS objects, which are related to the rotated object detection task. Nevertheless, the models after MTP demonstrate an excellent capability in detecting horizontal boxes.

\subsection{Rotated Object Detection}
We then investigate the impact of MTP on the rotated object detection task, which is a typical RS task distinguished from natural scene object detection because of special overhead views. This is also one of the motivations to implement MTP using SAMRS.

\begin{table*}[t]
  \caption{The mAP (\%) of finetuning different pretrained models on the DIOR-R, FAIR1M-2.0, DOTA-V1.0, and DOTA-V2.0 datasets. MS indicates whether the accuracy on DOTA-V1.0 is obtained from the multi-scale training and testing. $\dagger$: The feature pyramid is formed by upsampling and downsampling the last layer feature of the backbone network by following the strategy of ViTDet \cite{vitdet}.}
  \newcommand{\tabincell}[2]{\begin{tabular}{@{}#1@{}}#2\end{tabular}}
  \centering
  \begin{tabular}{llccccc}
  \hline
  \bfseries Method &\bfseries Backbone & \bfseries  MS &\bfseries  DIOR-R &\bfseries FAIR1M-2.0  & \bfseries DOTA-V1.0 & \bfseries DOTA-V2.0\\
  \hline
  RetinaNet-OBB \cite{retinanet} & ResNet-50 & - & 57.55 & -  & - &  46.68 \\
  Faster RCNN-OBB \cite{FasterRCNN} &  ResNet-50 & \ding{56} & 59.54 & - & 69.36 & 47.31 \\
  FCOS-OBB \cite{fcos} & ResNet-50 & -  & -  & -  & - &  48.51 \\
  ATSS-OBB \cite{atss} & ResNet-50 & \ding{56} & -  & -  & 72.29 & 49.57 \\
  SCRDet \cite{scrdet} & ResNet-101 & \ding{56} & - & - & 72.61 & - \\
  Gilding Vertex \cite{gliding_vertex} & ResNet-50 & \ding{56} & 60.06 & - & 75.02 & - \\
  ROI Transformer \cite{roi_transformer} & ResNet-50 &  \ding{52} & 63.87 & -  & 74.61 & 52.81 \\
  CACo \cite{caco} & ResNet-50 & -  & 64.10  &  47.83 & - & - \\
  RingMo \cite{ringmo} & Swin-B & - & -  & 46.21 & - & - \\
  R$^3$Det \cite{r3det} & ResNet-152 & \ding{52} & - &-  & 76.47 &- \\
  SASM \cite{sasm} & ResNeXt-101 &  \ding{52} & - & - & 79.17 & 44.53 \\
  AO2-DETR \cite{ao2_detr} & ResNet-50 & \ding{52}  & - & - & 79.22 & - \\
  S$^2$ANet \cite{aod_2022_tgrs_s2anet} & ResNet-50 & \ding{52} & - & - & 79.42 & 49.86 \\
  ReDet \cite{aod_2021_cvpr_redet} & ReResNet-50 & \ding{52} & -  &   & 80.10  &   \\
  R$^3$Det-KLD \cite{aod_2021_nips_kld}  &  ResNet-50 & \ding{52} & - & - & 80.17 & 47.26 \\
  R$^3$Det-GWD \cite{aod_2021_icml_gwd} & ResNet-152 & \ding{52} &  - & - & 80.23 & - \\
  R$^3$Det-DEA \cite{aod_2022_tgrs_dea} & ReResNet-50 &  \ding{52} & - & - & 80.37 & - \\
  AOPG \cite{aod_2022_tgrs_dior_r_aopg} & ResNet-50 & \ding{52} & 64.41 & - & 80.66 & - \\
  DODet \cite{aod_2022_tgrs_dodet} & ResNet-50 & \ding{52} & 65.10 & - & 80.62  & - \\
  PP-YOLOE-R \cite{pp_yoloe_r} & CRN-x \cite{pp_yoloe_crn} &  \ding{52} &  &  & 80.73 & \\
  GASSL \cite{gassl} & ResNet-50 & - &  65.65 & 48.15 & - & - \\
  SatMAE \cite{satmae} & ViT-L & - & 65.66  & 46.55 & - & - \\
  TOV \cite{tov} & ResNet-50 & - & 66.33 & 49.62 & - & - \\
  Oriented RepPoints \cite{aod_2022_cvpr_oriented_reppoint} & ResNet-50 & \ding{56} & 66.71 & - & 75.97 & 48.95 \\
  GGHL \cite{gghl} & DarkNet-53\cite{yolov3_darknet} &  \ding{56} & 66.48 & -  & 76.95 & 57.17 \\
  CMID \cite{cmid} & Swin-B & \ding{56} & 66.37 & 50.58  & 77.36 & - \\
  RSP \cite{rsp} & ViTAEv2-S & \ding{56}  & - & - & 77.72 & - \\
  Scale-MAE \cite{scale_mae} & ViT-L & -  & 66.47 & 48.31 & - & - \\
  SatLas \cite{satlas} & Swin-B & - & 67.59 & 46.19 & - & - \\
  GFM \cite{gfm} & Swin-B & - & 67.67 & 49.69 & - & - \\
  PIIDet \cite{piidet} & ResNeSt-101 & \ding{52} & 70.35 & -  & 80.21 & - \\
  Oriented RCNN \cite{orcn} & ResNet-50 & \ding{52} & - & - & 80.87 & 53.28 \\
  R$^3$Det-KFIoU \cite{kfiou} & ResNet-152 & \ding{52} & - & -  &  81.03 & - \\
  RTMDet-R \cite{rtmdet} & RTMDet-R-l &\ding{52}  & - & - & 81.33 & - \\
  DCFL \cite{dcfl} & ReResNet-101 \cite{aod_2021_cvpr_redet} & \ding{56}  & 71.03 & - & - & 57.66 \\
  SMLFR \cite{smlfr} & ConvNeXt-L \cite{convnext} & \ding{56} & 72.33  & -  & 79.33  & - \\
  ARC \cite{arc} & ARC-R50 & \ding{52} & - & - & 81.77\textcolor{green}{*} & - \\
  LSKNet \cite{lsknet} & LSKNet-S & \ding{52} & - & -  & 81.85\textcolor{red}{*} & - \\
  STD \cite{std} & ViT-B & \ding{52} & - & - & 81.66 &- \\
  STD \cite{std} &  HiViT-B \cite{hivit} & \ding{52} & - & - & 82.24* & -\\
  BillionFM \cite{bfm} & ViT-G12X4 & - & 73.62\textcolor{green}{*} & - & - & 58.69\textcolor{red}{*} \\
  SkySense \cite{skysense} & Swin-H &  - & 74.27\textcolor{red}{*}  & 54.57* & - & - \\
  RVSA \cite{rvsa} $\dagger$ & ViT-B + RVSA  & \ding{52} & 70.67 & - & 81.01 & - \\
  \hline
  MAE & ViT-B + RVSA & \ding{52} & 68.06 & 51.56  & \bfseries 80.83 & 55.22 \\
  MAE + MTP &  ViT-B + RVSA &  \ding{52} & \bfseries 71.29  &  \bfseries 51.92 & 80.67  & \bfseries 56.08 \\
  \hline
  MAE & ViT-L + RVSA & \ding{52}  & 70.54  & \bfseries 53.20\textcolor{red}{*} & 81.43 & \bfseries 58.96* \\
  MAE + MTP &   ViT-L + RVSA & \ding{52} & \bfseries 74.54*  & 53.00\textcolor{green}{*}  & \bfseries 81.66  & 58.41\textcolor{green}{*} \\
  \hline
  IMP & InternImage-XL & \ding{52} & 71.14 & 50.67 & 80.24 & 54.85 \\
  IMP + MTP & InternImage-XL & \ding{52}  & \bfseries 72.17 & \bfseries 50.93  &  \bfseries 80.77 & \bfseries 55.13  \\
  \hline
\end{tabular}
\label{rot_det}
\end{table*}

\subsubsection{Dataset}
We adopt the four most commonly used datasets for this task: DIOR-R \cite{aod_2022_tgrs_dior_r_aopg}, FAIR1M-2.0 \cite{fair1m}, DOTA-V1.0 \cite{dota1} and DOTA-V2.0 \cite{dotav2}.

\begin{enumerate}
\item[1)] DIOR-R: This is the extended oriented bounding box version of DIOR \cite{dior}. It has 23,463 images and 192,158 instances over 20 classes. Each image in this dataset has been cropped into 800 $\times$ 800. Following \cite{rvsa, bfm, skysense}, we merge the original training and validation sets for training, while the testing set is used for evaluation.
\item[2)] FAIR1M-2.0: This is a large-scale RS benchmark dataset, including more than 40,000 images and 1 million instances for fine-grained object detection. It collects samples with resolutions ranging from 0.3 to 0.8$m$ and image sizes ranging from 1,000 $\times$ 1,000 to 10,000 $\times$ 10,000 from various sensors and platforms. It contains 37 subcategories belonging to 5 classes: Ship, Vehicle, Airplane, Court, and Road. In this paper, we use the more challenging version of 2.0, which additionally incorporates the 2021 Gaofen Challenge dataset. The training and validation sets are together adopted for training.
\item[3)] DOTA-V1.0: This is the most popular dataset for RS rotated object detection. It comprises 2,806 images spanning from 800 $\times$ 800 to 4,000 $\times$ 4,000 $\times$, where 188,282 instances from 15 typical categories are presented. We adopt classical train/test split, that is, the original training and validation sets are together for training, while the original testing set is used for evaluation.
\item[4)] DOTA-V2.0: The is the enhanced version of DOTA V1.0. By additionally collecting larger images, adding new categories, and annotating tiny instances, it finally contains 11,268 images, 1,793,658 instances, and 18 categories. We use the combination of training and validation sets for training, while the test-dev set is used for evaluation.
\end{enumerate}

\subsubsection{Implementation Details} Since the large-size image is not suitable for training, we first perform data cropping. For DOTA-V2.0, we adopt single-scale training and testing by following \cite{dcfl}, where the images are cropped to patches in size of 1,024 $\times$ 1,024 with an overlap of 200. For DOTA-V1.0 and FAIR1M-2.0, we implement the multiscale training and testing, \textit{i.e.}, the original images are scaled with three ratios: (0.5, 1.0, 1.5). Then, the DOTA-V1.0 images are cropped to 1,024 $\times$ 1,024 patches but with an overlap of 500, while FAIR1M-2.0 images adopt a patch size of 800 and an overlap of 400. The batch sizes are set to 4, 16, 4, and 4 for the DIOR-R, FAIR1M, DOTA-V1.0, and DOTA-V2.0 datasets, respectively. The other settings during training are the same as horizontal object detection. We adopt the Oriented-RCNN network implemented in MMRotate. During training, input data is augmented by random flipping and random rotation. The mean average precision (mAP) is adopted as the evaluation metric.

\subsubsection{Finetuning Results and Analyses}

Table \ref{rot_det} shows the finetuning results. Except for DIOR-R, we find the MTP pretrained models cannot always demonstrate obvious advantages compared to their counterparts. Since the volumes of FAIR1M-2.0, DOTA-V1.0, and DOTA-V2.0 are much larger than DIOR-R, we speculate that after long-time finetuning, the benefit of MTP becomes diminished. We will further explore this issue in later sections. Nevertheless, owing to the excellent structure, RVSA-L outperforms the ViT-G-based foundation model \cite{bfm} with over 1 billion parameters on DOTA-V2.0. Compared to the powerful SkySense model \cite{skysense}, our models achieve better performance on the DIOR-R. While on FAIR1M-2.0, except SkySense, our models surpass all other methods by a large margin. Generally, our models have comparable representation capability as SkySense, although it has over 600M parameters and utilizes 20 million images for pretraining. We also notice the performances of our models still have gaps compared with the current advanced method STD \cite{std} on DOTA-V1.0. It may be attributed to the adopted classical detector Oriented-RCNN \cite{orcn}, which limits the detection performance.

\subsection{Semantic Segmentation}
We further consider finetuning the pretrained models on the finer pixel-level tasks, \textit{e.g.}, the semantic segmentation task. It is one of the most important RS applications for the extraction and recognition of RS objects and land covers.

\subsubsection{Dataset} We separately take into account both single-class geospatial target extraction and multi-class surface element perception through two RS semantic segmentation datasets: SpaceNetv1 \cite{spacenet} and LoveDA \cite{loveda}. Here are their details:
\begin{enumerate}
    \item[1)] SpaceNetv1: This dataset is provided by SpaceNet Challenge \cite{spacenet} for extracting building footprints. It is made up of the DigitalGlobe WorldView-2 satellite imagery with a ground sample distance of 0.5$m$ photoed during 2011-2014 over Rio de Janeiro. It covers about 2,544 $km^2$, including 382,534 building instances. Since only the 6,940 images in the original training set are available, following \cite{satmae, ctxmim}, we randomly split these images into two parts, where the first part containing 5,000 images being used as the training set, and another part will be used for testing.
    \item[2)] LoveDA: This is a challenging dataset involving both urban and rural scenes. It collects 0.3$m$ spaceborne imagery from Google Earth, where the images were obtained in July 2016, covering 536.15$km^2$ of Nanjing, Changzhou, and Wuhan. It has 5,987 images in size of 1,024 $\times$ 1,024, involving seven types of common land covers. We merge the official training and validation sets for training and conduct evaluation using the official testing set. 
\end{enumerate}

\begin{table}[t]
  \caption{The mIOU (\%) of finetuning different pretrained models with UperNet on the SpaceNetv1 and LoveDA datasets.}
  \newcommand{\tabincell}[2]{\begin{tabular}{@{}#1@{}}#2\end{tabular}}
  \centering
  \begin{tabular}{llccc}
  \hline
   \bfseries Method & \bfseries Backbone & \bfseries SpaceNetv1 &\bfseries  LoveDA \\
  \hline
  PSANet \cite{psanet} & ResNet-50  & 75.61 & - \\
  SeCo \cite{seco} &  ResNet-50 & 77.09  & 43.63  \\
  GASSL \cite{gassl} & ResNet-50 &  78.51 & 48.76 \\
  SatMAE \cite{satmae} & ViT-L & 78.07 & - \\
  CACo \cite{caco} & ResNet-50  &  77.94 & 48.89 \\
  PSPNet \cite{pspnet} & ResNet-50 & - & 48.31 \\
  DeeplabV3+ \cite{deeplabv3_p} & ResNet-50 & - & 48.31 \\
  FarSeg \cite{farseg} & ResNet-50 & - & 48.15 \\
  FactSeg \cite{ass_2022_tgrs_factseg} & ResNet-50 & - & 48.94 \\
  TOV \cite{tov} &  ResNet-50 & - & 49.70 \\
  HRNet  \cite{hrnet} & HRNet-W32 & - & 49.79 \\
  GeRSP \cite{gersp}  & ResNet-50 & - & 50.56 \\
  DCFAM \cite{dcfam} & Swin-T & - & 50.60 \\
  UNetFormer \cite{unetformer} & ResNet-18 & - & 52.40 \\
  RSSFormer \cite{rssformer} & RSS-B & - & 52.43 \\
  UperNet \cite{upernet} & ViTAE-B + RVSA \cite{rvsa} & - & 52.44 \\
  Hi-ResNet \cite{hiresnet} & Hi-ResNet & - & 52.50 \\
  RSP \cite{rsp} & ViTAEv2-S & - & 53.02 \\
  SMLFR \cite{smlfr} & ConvNext-L & - & 53.03 \\
  LSKNet \cite{lsknet} & LSKNet-S & - & 54.00 \\
  CtxMIM \cite{ctxmim}  & Swin-B & 79.47  & - \\
  AerialFormer \cite{aerialformer} & Swin-B  & - & 54.10 \\
  BillionFM \cite{bfm} & ViT-G12X4 & - & 54.40* \\
  \hline
  MAE & ViT-B + RVSA &  79.56\textcolor{green}{*} &  51.95 \\
  MAE + MTP &  ViT-B + RVSA  & \bfseries 79.63\textcolor{red}{*}  & \bfseries 52.39 \\
  \hline
  MAE & ViT-L + RVSA  &  \bfseries 79.69*   & 53.72\\
  MAE + MTP &   ViT-L + RVSA & 79.54  & \bfseries 54.17\textcolor{red}{*} \\
  \hline
  IMP & InternImage-XL  & 79.08 & 53.93\textcolor{green}{*} \\
  IMP + MTP & InternImage-XL  & \bfseries 79.16 & \bfseries 54.17\textcolor{red}{*} \\
  \hline
\end{tabular}
\label{sem_seg}
\end{table}

\subsubsection{Implementation Details} Except that the models are trained with 80K iterations with a batch size of 8, and the warming up stage in the parameter scheduler lasts 1,500 iterations, most of the optimization settings are similar to the scene classification section. We use the UperNet \cite{upernet} as the segmentation framework, where the input image sizes during training are 384 $\times$ 384 and 512 $\times$ 512 for SpaceNetv1 and LoveDA, respectively, through random scaling and cropping. We also adopt random flipping for data augmentation. All experiments are implemented by MMSegmentation, where the mean value of the intersection over union (mIOU) is adopted as the evaluation metric.

\subsubsection{Finetuning Results and Analyses}

The results presented in Table \ref{sem_seg} demonstrate that MTP is also useful for enhancing the models' performance on semantic segmentation tasks. Compared to SpaceNetv1, the improvements on the classical land cover classification dataset: LoveDA, are even more significant. As a result, on this dataset, our models surpass all previous methods except the BillionFM \cite{bfm}, which utilizes a model with over 1 billion parameters. On the SpaceNetv1, our models set new state-of-the-art accuracy. Nonetheless, probably due to overfitting, the results of SpaceNetv1 also indicate that the performances on simple extraction tasks do not improve as increasing model capacity. We have also noticed the performance of ViT-L + RVSA on SpaceNetv1 is decreased when adopting MTP. We will conduct further exploration in later sections.

\subsection{Change Detection}

\begin{table*}[t]
  \caption{The F1 score (\%) of finetuning different pretrained models with UNet on the OSCD, WHU, LEVIR, and SVCD/CDD datasets.}
 \centering
 \resizebox{0.65\linewidth}{!}{
 \begin{tabular}{lccccc}
  \hline
  \bfseries Method & \bfseries Backbone & \bfseries OSCD & \bfseries WHU & \bfseries LEVIR & \bfseries SVCD/CDD \\
  \hline
  GASSL \cite{gassl} & ResNet-50 & 46.26 &- & -  & - \\
  SeCo \cite{seco} & ResNet-50 & 47.67 & - & 90.14 & - \\
  FC-EF \cite{fcsn_cd} & - & 48.89 & - & 62.32 & 77.11\\
  SwiMDiff \cite{swimdiff} & ResNet-18 & 49.60 & - & - & - \\
  CACo \cite{caco} & ResNet-50 & 52.11 & - & - & - \\
  SatMAE \cite{satmae} &  ViT-L & 52.76  & - & - & - \\
  SNUNet \cite{acd_2021_grsl_snunet} & -  &- & 83.49 & 88.16 & 96.20 \\
  BIT \cite{acd_2021_tgrs_bit} & ResNet-18  &  - & 83.98  & 89.31 & - \\
  SRCDNet \cite{acd_2021_tgrs_srcdnet} & ResNet-18 & - & 87.40 & - & 92.94 \\
  CLNet \cite{acd_2021_isprs_clnet} & - & - & - & 90.00 & 92.10 \\
  HANet \cite{hanet} & - & - & - & 90.28 & - \\
  RECM \cite{recm} & ViT-S & - & - & 90.39 & - \\
  ChangeFormer \cite{changeformer} & MiT-B2 \cite{segformer} & - & - & 90.40 & - \\
  AERNet \cite{aernet} & ResNet-34 & - & - & 90.78  & - \\
  ESCNet  \cite{escnet} & - & - & - & - & 93.54 \\
  DSAMNet \cite{acd_2021_tgrs_dsamnet} & ResNet-18 & - & - & - & 93.69 \\
  GCD-DDPM \cite{gcdddpm} & - & - & 92.54 & 90.96 & 94.93 \\
  CDContrast \cite{cdcontrast}& - & - & - & -  & 95.11 \\
  DDPM-CD \cite{ddpmcd} & UNet \cite{unet} & - & 92.65 & 90.91 & 95.62 \\
  DeepCL \cite{deepcl} & EfficientNet-b0 \cite{efficientnet} & - & - & 91.11 & - \\
  DMNet \cite{dmnet} & ResNet-50 & - & - & - & 95.93 \\
  ChangeStar \cite{changestar} & ResNeXt-101 \cite{resnext}  & - & - & 91.25 & - \\
  RSP \cite{rsp}  & ViTAEv2-S \cite{vitae_v2} & - & - & 90.93 & 96.81 \\
  SAAN \cite{saan} & ResNet-18 & - & - & 91.41 & 97.03 \\
  SiamixFormer \cite{siamixformer} & MiT-B5 \cite{segformer} & - & - & 91.58 & 97.13 \\
  TransUNetCD \cite{transunetcd} & ResNet-50  &  - & 93.59 & 91.11 & 97.17 \\
  RDPNet  \cite{rdpnet} & - & - & - & 90.10 & 97.20 \\
  SDACD \cite{sdacd} & - & - & - & - & 97.34 \\
  Siam-NestedUNet \cite{siamesenestedunet} & UNet++ \cite{unet++} & - & - & 91.50 & - \\
  Changen \cite{changen} & MiT-B1 \cite{segformer} & - & - & 91.50 & - \\
  HCGMNet \cite{hcgmnet} & VGG-16  & - & - & 91.77 & - \\
  CEECNet \cite{ceecnet} & - & - & - & 91.83 & - \\
  RingMo \cite{ringmo}  & Swin-B & - & - & 91.86 & - \\
  CGNet \cite{cgnet} & VGG-16 & - & - & 92.01 & - \\
  TTP \cite{ttp} & SAM \cite{sam} & - & - & 92.10 & - \\
  Changer \cite{changer} & ResNeSt-101 \cite{resnest} & - & - & 92.33 &- \\
  WNet \cite{wnet} & ResNet-18 + DAT \cite{dat} &  - & 91.25 & 90.67 & 97.56 \\
  SpectralGPT$^+$ \cite{spectralgpt} & ViT-B & 54.29 & - & - & - \\
  C2FNet \cite{c2fnet} & VGG-16 & - & - & 91.83 & - \\
  MATTER \cite{matter} & ResNet-34  & 59.37\textcolor{green}{*} & -  & - &  - \\
  GFM \cite{gfm} & Swin-B  & 59.82\textcolor{red}{*} & - & - & - \\
  FMCD \cite{fmcd} & EfficientNet-b4 \cite{efficientnet} & - & 94.48 & - & - \\
  SGSLN/256 \cite{sgsln} & - & - & 94.67 & 91.93  & 96.24 \\
  P2V-CD \cite{p2vcd} & - & - & 92.38 & 91.94 & 98.42* \\
  ChangeCLIP \cite{changeclip} & CLIP \cite{clip} & - & 94.82\textcolor{green}{*} & 92.01 & 97.89 \\
  BAN \cite{ban_cd} & InternImage-XL \cite{internimage} & -  & - & 91.94 & - \\
  BAN \cite{ban_cd} & ViT-L \cite{internimage} & -  & - & 91.96 & - \\
  BAN \cite{ban_cd} & ChangeFormer \cite{changeformer} & -  & - & 92.30 & - \\
  SkySense \cite{skysense} & Swin-H  & 60.06* & - & 92.58\textcolor{red}{*} & - \\
  \hline
   MAE &  ViT-B + RVSA & 50.28  & 93.77  & 92.21 & 97.80 \\
   MAE + MTP &  ViT-B + RVSA  & \bfseries 53.36 &\bfseries 94.32  &\bfseries  92.22 & \bfseries 97.87 \\
  \hline
  MAE &  ViT-L + RVSA & 54.04 & 94.07 & 92.52 &  97.78 \\
  MAE + MTP  &  ViT-L + RVSA & \bfseries 55.92 & \bfseries 94.75  & \bfseries 92.67* & \bfseries 97.98  \\
  \hline 
 IMP &  InternImage-XL & 51.61 & 95.33\textcolor{red}{*} & 92.46 &  \bfseries 98.37\textcolor{red}{*} \\
 IMP + MTP & InternImage-XL & \bfseries 55.61 & \bfseries 95.59* & \bfseries 92.54\textcolor{green}{*} & 98.33\textcolor{green}{*} \\
  \hline
\end{tabular}
}
\label{change_detection}
\end{table*}

Finally, we pay attention to the change detection task, which can be regarded as a special type of segmentation by extracting the changed area between the RS images taken at different times in the same location. Here, we mainly consider the most representative bi-temporal change detection. 

\subsubsection{Dataset} We conduct the finetuning on the datasets of different scales: Onera Satellite Change Detection Dataset (OSCD) \cite{oscd}, Wuhan University Building Change Detection Dataset (WHU) \cite{whu_cd}, the Learning, Vision, and Remote Sensing Change Detection Dataset (LEVIR) \cite{levir}, and the Season-Varying Change Detection Dataset (SVCD) \cite{cdd}, which is also called ``CDD''.
\begin{enumerate}
    \item[1)] OSCD: This is a small-scale dataset. It contains 24 pairs of Sentinel-2 multispectral images involving all bands and in an average size of 600 $\times$ 600. These images are obtained during 2015-2018 to record urban changes. We follow the same train/val split as \cite{oscd}, where training and validation sets include 14 and 10 pairs, respectively.
    \item[2)] WHU: This dataset is used for detecting building changes in a single view. It contains two large-scale images with a ground resolution of 0.3$m$ and in size of 32,507 $\times$ 15,354. They are collected in 2012 and 2016, containing 12,796 and 16,077 instances, respectively. Since there is no official data split, the 70\%, 10\%, and 20\% patches of the cropped images are randomly selected as training, validation, and testing sets as suggested by \cite{sgsln}.
    \item[3)] LEVIR: This dataset contains 637 pairs of 1,024 $\times$ 1,024 images with a spatial resolution of 0.5$m$. These images are acquired between 2002 and 2018 from 20 different regions in Texas, USA. It contains 31,333 change instances. We adopt the official split, where training, validation, and testing sets contain 445, 64, and 128 pairs, respectively.
    \item[4)] SVCD/CDD: This dataset focuses on seasonal variations. It initially contains 11 pairs of images obtained from Google Earth in different seasons, with spatial resolutions ranging from 0.03 to 1$m$. It now has been cropped to 16,000 pairs of patches in size of 256 $\times$ 256 by \cite{cdd_clip}. The 10,000/3,000/3,000 pairs are separately used as training, validation, and testing sets.
\end{enumerate}

\subsubsection{Implementation Details}
Following \cite{seco, skysense}, we crop the OSCD images to 96 $\times$ 96 patches with no overlapping, obtaining 827/385 pairs for training/testing. However, the training is difficult to converge due to the extremely small input size, thus we rescale the image to 224 $\times$ 224 before inputting it into the network. For the WHU dataset, we separately have 5334, 762, and 1524 images for training, validation, and testing, after cropping the image to patches in size of 256 $\times$ 256 without overlaps. A similar operation is conducted for LEVIR, generating training, validation, and testing sets containing 7120, 1024, and 2048 samples, respectively. The training epochs on OSCD, WHU, LEVIR, and CDD are separately set to 100, 200, 150, and 200. The batch size of all datasets is uniformly set to 32. We adopt the same optimization strategy as the scene classification task. To fully leverage the feature pyramid produced by foundation models, we adopt a UNet \cite{unet} to process the differences between different temporal features. The training is implemented through Open-CD\footnote{https://github.com/likyoo/open-cd}, where the data augmentation includes random rotation, random flipping, random exchange temporal, and color jitters that randomly adjust brightness, contrast, hue, and saturation of images. The F1 score of the changed class is adopted as the evaluation metric.

\subsubsection{Finetuning Results and Analyses}

\begin{table*}[t]
  \caption{Detailed hyperparameter settings in finetuning pretrained models on different datasets. ``\ding{52}'' and ``\ding{56}'' indicate whether the MTP is useful for improving performance compared to the setting without MTP.}
  \centering
  \resizebox{\linewidth}{!}{
  \begin{tabular}{l|cccccccc}
  \hline
   \bfseries\multirow{2}{*}{\normalsize{Dataset}}  & \multicolumn{2}{c}{\normalsize{\textbf{\textit{\textcolor[rgb]{0,0.8,0.5}{Scene Classification}}}} }  &  \multicolumn{2}{c}{\normalsize{\textbf{\textit{\textcolor[rgb]{1,0.5,0}{Horizontal Detection}}}}}   & \multicolumn{4}{c}{\normalsize{\textbf{\textit{\textcolor[rgb]{1,0,0}{Rotated Object Detection}}}}}  \\
   & \ EuroSAT &  RESISC-45 &  Xview  &  DIOR  &  DIOR-R  &  FAIR1M-2.0 &  DOTA-V1.0 &  DOTA-2.0  \\
  \hline 
  Training Image Number ($N_{TrIm}$) & 16,200 & 6,300 & 20,084 & 11,725 & 11,725 & 288,428 & 133,883 &  31,273\\
  Training Epoch Number ($N_{TrEp}$) & 100 & 200  & 12 & 12 &  12 & 12 &  12 & 40 \\
  Total Sample Number ($N_{ToSa}$) & 1,620,000 & 1,260,000 &241,008  & 140,700 & 140,700 & 3,461,136  & 1,606,596 & 1,250,920  \\
  Batch Size ($S_B$) & 64 & 64 & 8 & 4 & 4 & 16 & 4 &4  \\
  Total Iteration Number ($N_{ToIt}$)  & 25,312 & 19,688 & 30,126 & 35,175 & 35,175 & 216,321 & 401,649  & 312,730 \\
  Training Image Size ($S_{TrIm}$) & 224 & 224 & 416 & 800 & 800 & 800 & 1,024 & 1,024 \\
  Class Number ($N_C$) & 10 & 45 & 60 & 20 & 20 & 37 & 15 & 18 \\
  \hline
  Average Pixel per Class ($AP_C$)  & 36,288,000  & 6,272,000 &167,0988 & 5,628,000  & 5,628,000 &  74,835,373 & 109,676,954 & 71,163,449 \\
  Average Iteration per Class ($AI_C$)  & 2,531  & 438 & 502 & 1759 & 1,759  & 5,847  & 26,777 & 17,374 \\
  ViT-B + RVSA &\ding{52}  & \ding{52} & \ding{52} & \ding{52} &\ding{52} & \ding{52} &  \ding{56} &\ding{52}  \\
  ViT-L + RVSA & \ding{52}  & \ding{52} & \ding{52} & \ding{52} & \ding{52} & \ding{56}&\ding{52} &\ding{56} \\
  InternImage-XL& \ding{56}  & \ding{52} & \ding{52} & \ding{52} & \ding{52} &\ding{52} & \ding{52} & \ding{52}\\
  \hline
   \bfseries\multirow{2}{*}{\normalsize{Dataset}}  &  \multicolumn{2}{c}{\normalsize{\textbf{\textit{\textcolor[rgb]{0,0.5,1}{Semantic Segmentation}}}}}   & \multicolumn{4}{c}{\normalsize{\textbf{\textit{\textcolor[rgb]{0.8,0.2,0.8}{Bi-temporal Change Detection}}}}} & & \\
   &  SpaceNetv1 &  LoveDA &  OSCD  &  WHU  &  LEVIR  &   SVCD/CDD &  &  \\
  \hline  
  Training Image Number ($N_{TrIm}$)  & 5,000 & 4,191 & 827 &  5,334 & 7,120 & 10,000  & &  \\
  Training Epoch Number ($N_{TrEp}$) & 128  & 153 & 100 & 200 & 150 & 200  & &  \\
  Total Sample Number ($N_{ToSa}$) & 640,000 & 640,000 & 82,700 & 106,6800 & 1,068,000 &2,000,000 & &  \\
  Batch Size ($S_B$)   & 8 & 8 & 32 & 32 & 32 & 32 & &  \\
  Total Iteration Number ($N_{ToIt}$)  & 80,000 & 80,000 & 2,584 & 33,338  & 33,375 &62,500  & &  \\
  Training Image Size ($S_{TrIm}$)  & 384 & 512 & 224 & 256 & 256 & 256 & &  \\
  Class Number ($N_C$)  & 2 & 7 & 2 & 2 & 2 & 2 & &  \\
  \hline
 Average Pixel per Class ($AP_C$)  & 122,880,000  &  46,811,429 &  9,262,400 & 136,550,400  & 136,704,000  & 256,000,000  & &  \\
  Average Iteration per Class ($AI_C$) & 40,000 & 11,429 & 1,292 &16,669  &  16,688 &  31,250  & &  \\
  ViT-B + RVSA  & \ding{52} & \ding{52} & \ding{52} & \ding{52} &\ding{52} & \ding{52} &   \\
  ViT-L + RVSA   &\ding{56}  &  \ding{52} & \ding{52} & \ding{52} & \ding{52}& \ding{52} &  \\
  InternImage-XL  & \ding{52} & \ding{52} & \ding{52} &  \ding{52}&\ding{52} & \ding{56} &   \\
  \hline
\end{tabular}
 }
 \label{influence_factor}
\end{table*}

To comprehensively assess the finetuning performance of pretrained models, we conduct the comparison by collecting existing advanced change detection methods, as shown in Table \ref{change_detection}. It should be noted that, since the original WHU dataset does not provide an official train/test split, various split strategies are adopted in different methods. Therefore, on this dataset, we only list the accuracy value of the methods that employ the same settings as us or training with more images. It can be seen that MTP effectively improves the performances of pretrained models on these datasets. Especially, our models perform well on three large-scale datasets: WHU, LEVIR, and SVCD/CDD. Even if adopting simple UNet \cite{unet} and the RVSA model of the base version, the finetuning performances have been competitive and surpassed many advanced approaches. When utilizing larger models, the performance can be further boosted. Finally, they achieve the best accuracy on the WHU and LEVIR datasets by outperforming almost all existing methods, including the recent SkySense \cite{skysense} that builds a larger change detection network with over 600M parameters, ChangeCLIP \cite{changeclip} that uses CLIP \cite{clip} to obtain additional knowledge from language modalities, and the newly proposed adapter BAN \cite{ban_cd}, where the ability of existing foundation model and change detection approaches can be exploited. Different from large-scale scenes, on the small-scale dataset OSCD, although MTP is still useful, the performances of our models have relatively large gaps compared to current works. We attribute the reason to data discrepancy and image size. Specifically, our models are pretrained on high-resolution RS images, which are similar to another three change detection datasets. However OSCD images are captured by multispectral sensors with lower resolutions. In addition, OSCD images are only cropped to 96 $\times$ 96 during training, which may be unsuitable for feature extraction, especially for non-hierarchical vision transformers. In the comparison methods, MATTER \cite{matter}, SkySense \cite{skysense}, and GFM \cite{gfm} use pyramid feature networks. Among them, MATTER and SkySense adopt Sentinel-2 multispectral image in pretraining, while GFM crops the OSCD image to a larger size, i.e., 192 $\times$ 192. In contrast, a relatively small image size (128 $\times$ 128) restricts the performances of ViT in SpectralGPT \cite{spectralgpt}. These results suggest that it is necessary to conduct further explorations to enhance the model finetuning performance on out-of-domain datasets with small volumes and input sizes.

\subsection{Further Investigations and Analyses}

Besides evaluating the performances of pretrained models, we conduct further investigations to obtain deeper insights into the characteristics of MTP, including the influence factors of MTP, finetuning with fewer samples, and parameter reusing of decoders.

\subsubsection{Influence Factors of Multi-Task Pretraining}
Up to now, to comprehensively assess the impact of MTP, we have finetuned three types of foundation models, on five RS downstream tasks, involving a total of fourteen datasets. From the finetuning results (Table \ref{scene_cls}-\ref{change_detection}) we find that MTP improves these foundation models in most cases. But there are still some datasets, on which MTP does not perform well as expected, \textit{i.e.}, not all accuracies of three models are increased. To figure out the reason, we explore the influence factors related to the performance of MTP, as shown in Table \ref{influence_factor}. Intuitively, we suppose MTP may be affected by the characteristics of finetuning datasets and consider a series of variables, including ``Training Image Number'' ($N_{TrIm}$), ``Training Epoch Number'' ($N_{TrEp}$), ``Batch Size'' ($S_B$), and `` Training Image Size'' ($S_{TrIm}$). The ``Training Image Number'' means: for each dataset, the number of images used for training. For example, the $N_{TrIm}$ of DIOR is 11,725 since the original training and validation sets are together used for training. While ``Training Image Size'' represents the image size after data augmentation and preprocessing. Theoretically, we have
\begin{equation}
    N_{ToIt} = \frac{N_{TrIm} \cdot N_{TrEp}}{S_B},
\end{equation}
where $N_{ToIt}$ means the number of training iterations for model parameter updating under the mini-batch optimization strategy. In Table \ref{influence_factor}, we observe that as $N_{ToIt}$ increases, there is a tendency for MTP to have a negative impact on finetuning performance for a given task. However, this trend isn't universally applicable, as evidenced by varying results among pretrained models in segmentation tasks, all with the same $N_{ToIt}$. Additionally, we account for dataset difficulty by considering the Number of Classes $N_C$ and use the ``Average Iteration per Class'' ($AI_C$) to represent each dataset as
\begin{equation}
    AI_C = \frac{N_{ToIt}}{N_C}.
\end{equation}
Surprisingly, Table \ref{influence_factor} reveals a notable trend: a relatively large $AI_C$ corresponds to a negative impact of MTP for the same task. This suggests that, over extended finetuning periods, MTP models lose their advantage compared to conventional pretrained models. We propose a bold conjecture regarding this internal mechanism: \textit{the benefits of MTP diminish gradually due to excessive network optimization}. This discovery prompts a reconsideration of the trade-off between longer training times for accuracy gains and the benefits of pretraining when finetuning models. However, determining the critical point of $AI_C$ remains challenging due to limited experimentation, necessitating further investigation. It is important to note that this phenomenon differs from overfitting, as our models continue to outperform existing methods at this stage.

In addition to training duration, we consider dataset capacity, introducing the index ``Average Pixels per Class'' $AP_C$, which can be formulated by
\begin{equation}
    AP_C = \frac{N_{ToSa} \cdot S_{TrIm}}{N_C},
\end{equation}
where $N_{ToSa}=N_{TrIm} \cdot N_{TrEp}$ denotes the quantity of images processed during training. Consequently, $AP_C$ approximately reflects the data volume encountered during finetuning. Table \ref{influence_factor} reveals that $AP_C$ exhibits similar trends to $AI_C$, yet the correlation between $AP_C$ and MTP performance is less discernible compared to $AI_C$, possibly due to the presence of redundant pixels in RS images.

\begin{figure}[t]
    \centering
    \subfigure[]{\includegraphics[width=0.8\linewidth]{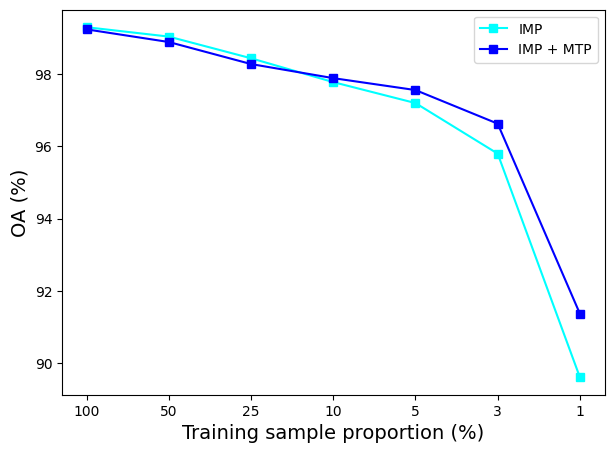}}\\
    \subfigure[]{\includegraphics[width=0.8\linewidth]{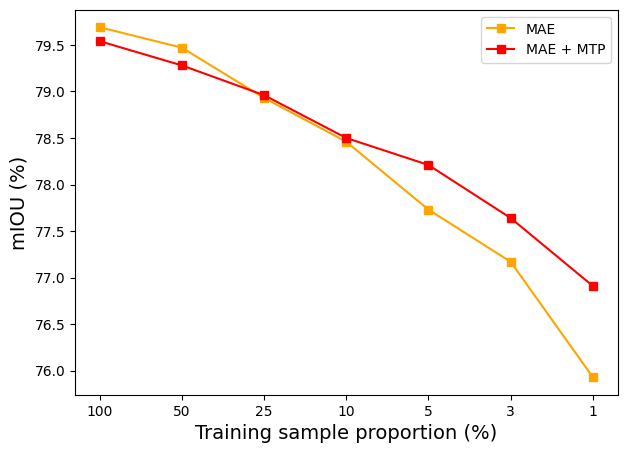}}
    \caption{The finetuning accuracy of different pretrained models with varying training sample sizes. (a) InternImage-XL on EuroSAT. (b) ViT-L + RVSA on SpaceNetv1.}
    \label{fewer_sample}
\end{figure}

\subsubsection{Fewer Sample Finetuning} 

The efficacy of SEP has been demonstrated in scenarios with limited samples \cite{samrs}. While MTP represents an extension of SEP, it is reasonable to anticipate that MTP could excel in analogous contexts. Moreover, as noted earlier, MTP primarily addresses the discrepancy between upstream pretraining and downstream finetuning tasks. This encourages us to consider that fewer downstream training samples might better showcase MTP's efficacy in facilitating efficient transfer from pretraining models. To explore this, we finetune InterImage-XL on EuroSAT and ViT-L + RVSA on SpaceNetv1, respectively, progressively reducing training samples. The results are depicted in Figure \ref{fewer_sample}. Initially, MTP's performance is slightly inferior to its counterparts when the training sample proportion is 100\%, as illustrated in Tables \ref{scene_cls} and \ref{sem_seg}. However, as training samples decrease, the performance curves converge until the training sample proportion is 10\%, at which point MTP's impact is minimal. Subsequent reductions in training samples lead to decreased accuracies across all models, yet the distances between the curves progressively widen. This trend suggests that the benefits of MTP are beginning to emerge, becoming increasingly significant. These findings validate our hypotheses, underscoring the benefit of MTP for finetuning foundational models on limited training samples.

\begin{figure*}[htbp]
  \centering
  \includegraphics[width=\textwidth]{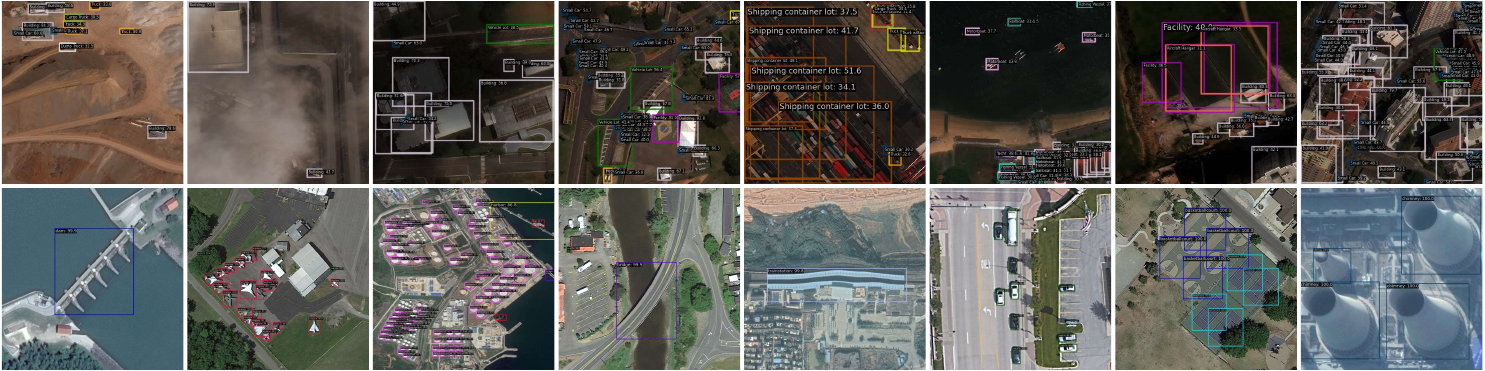}
  \caption{
  Visualization of the horizontal object detection predictions of MAE + MTP pretrained ViT-L + RVSA. The images of the first and the second rows are from Xview and DIOR testing sets, respectively.
  }
  \label{hdet_predicts}
\end{figure*}

\begin{figure*}[htbp]
  \centering
  \includegraphics[width=\textwidth]{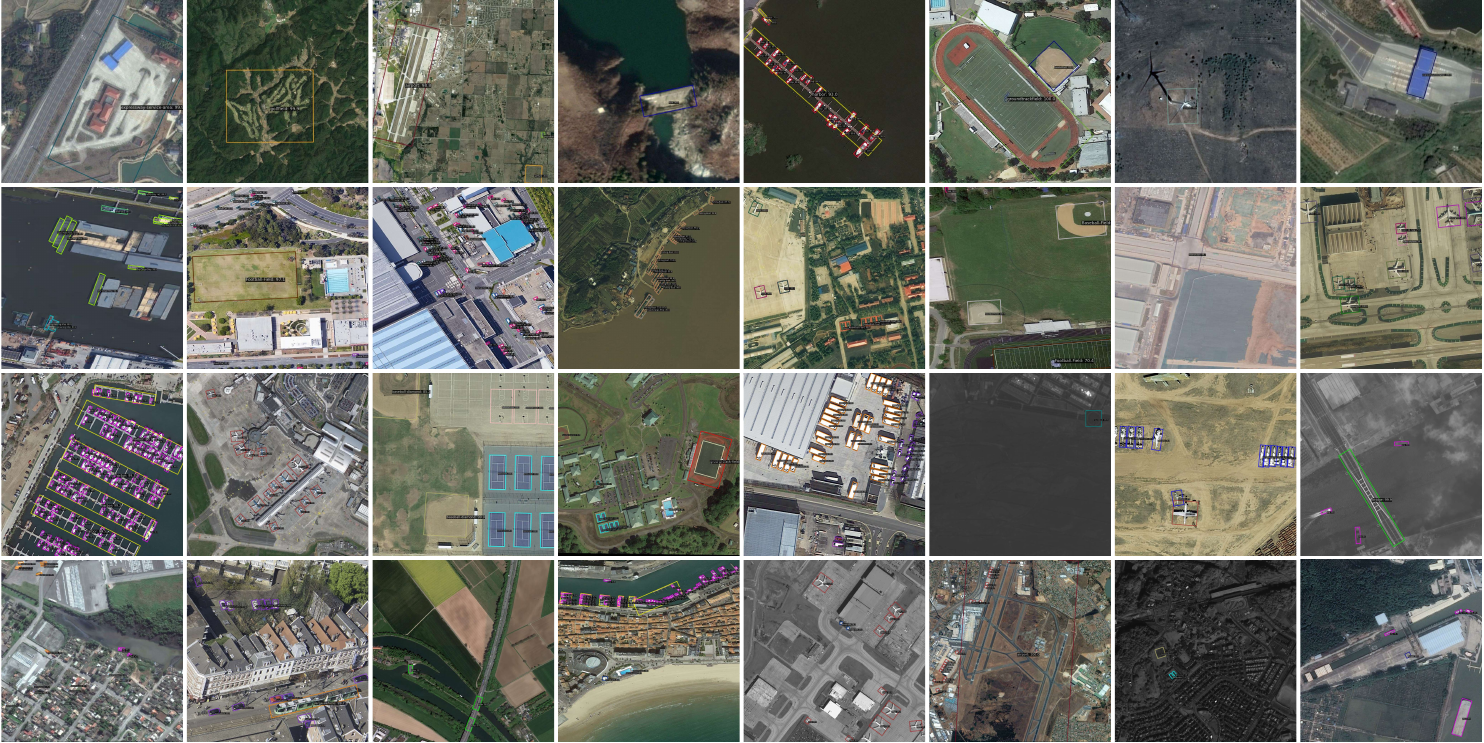}
  \caption{
  Visualization of the rotated object detection predictions of MAE + MTP pretrained ViT-L + RVSA. The images in four rows are from the testing sets of DIOR-R, FAIR1M-2.0, DOTA-V1.0 and DOTA-V2.0, respectively.
  }
  \label{rdet_predicts}
\end{figure*}

\begin{figure*}[htbp]
  \centering
  \includegraphics[width=\textwidth]{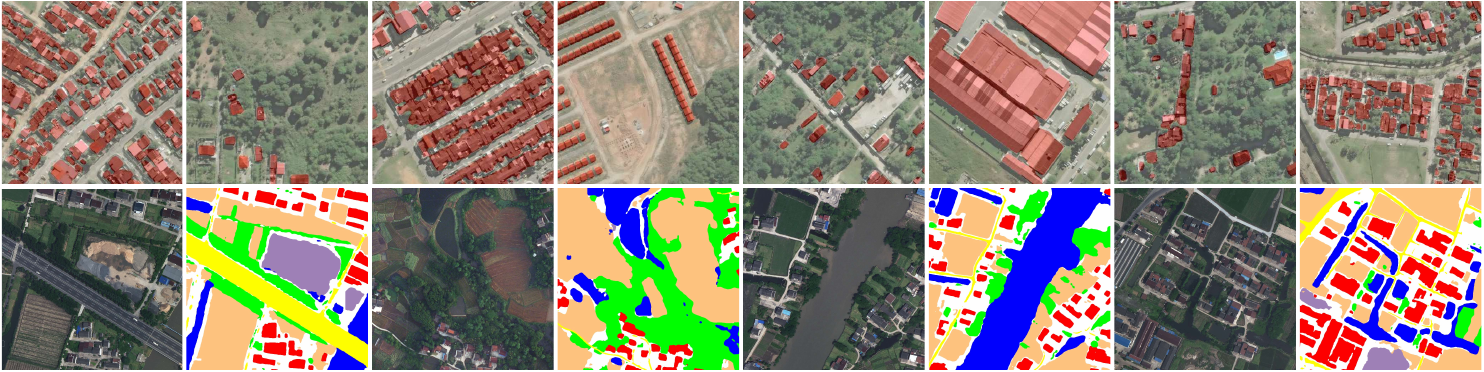}
  \caption{
  Visualization of the semantic segmentation predictions of MAE + MTP pretrained ViT-L + RVSA. The samples of the first and the second rows are from SpaceNetv1 and LoveDA testing sets, respectively.
  }
  \label{sseg_predicts}
\end{figure*}

\subsubsection{Decoder Parameter Reusing}

\begin{table}[htbp]
  \caption{
  The accuracies of finetuning ViT-B + RVSA on various datasets with and without reusing pretrained decoder weights.
  }
  \centering
  \begin{tabular}{c|ccc}
  \hline
   \bfseries Model  & \bfseries SpaceNetv1 & \bfseries LoveDA & \bfseries DIOR-R  \\
  \hline 
  w/o DPR & \bfseries 79.63  & \bfseries 52.39 & 71.29 \\
  w DPR & 79.54 & 51.83   & \bfseries 71.94 \\
  \hline
   \bfseries Model  & \bfseries FAIR1M-2.0  & \bfseries  DOTA-V1.0  & \bfseries DOTA-V2.0  \\
  \hline 
  w/o DPR & 51.92  & \bfseries 80.67 & 55.22 \\
  w DPR & \bfseries 52.19  & 80.54 & \bfseries 55.78 \\
  \hline
\end{tabular}
  \label{parameter_reuse}
\end{table}

MTP utilizes task-specific decoders for segmentation and detection tasks. Hence, reusing these decoder weights during finetuning seems a natural choice, and we conduct experiments accordingly using a backbone of ViT-B + RVSA. Specifically, aside from the backbone network, we also initialize the corresponding decoders with pretrained weights during finetuning. However, only semantic segmentation and rotated detection decoders are eligible for reuse, as per the segmentor or detector used in existing methods. Therefore, we performed the experiment on the corresponding six datasets, and the results have been presented in Table \ref{parameter_reuse}. Among four detection datasets, decoder parameter reusing (DPR) proves beneficial in three scenarios, which are actually employed in performing MTP. Nevertheless, on another classical dataset DOTA-V1.0, which can be regarded as a subset of DOTA-V2.0, DPR is insufficient useful. On segmentation tasks, the performances of DPR models are decreased. We speculate that besides the domain differences compared to pretraining datasets, current models may be additionally affected by the segmentation labels in pretraining. This is because decoders typically encode task-specific information. However, given that the SAMRS dataset used for pretraining involves annotations generated by SAM \cite{sam}, they inevitably contain errors, jeopardizing the quality of pretrained decoders. Finally, based on the above considerations, we conclude that \textit{after MTP, the performances of reusing pretrained decoder parameters in finetuning may depend on the similarity between pretraining and finetuning scenarios and the quality of pretraining annotations.}

\subsection{Visualization}

\begin{figure*}[htbp]
  \centering
  \includegraphics[width=\textwidth]{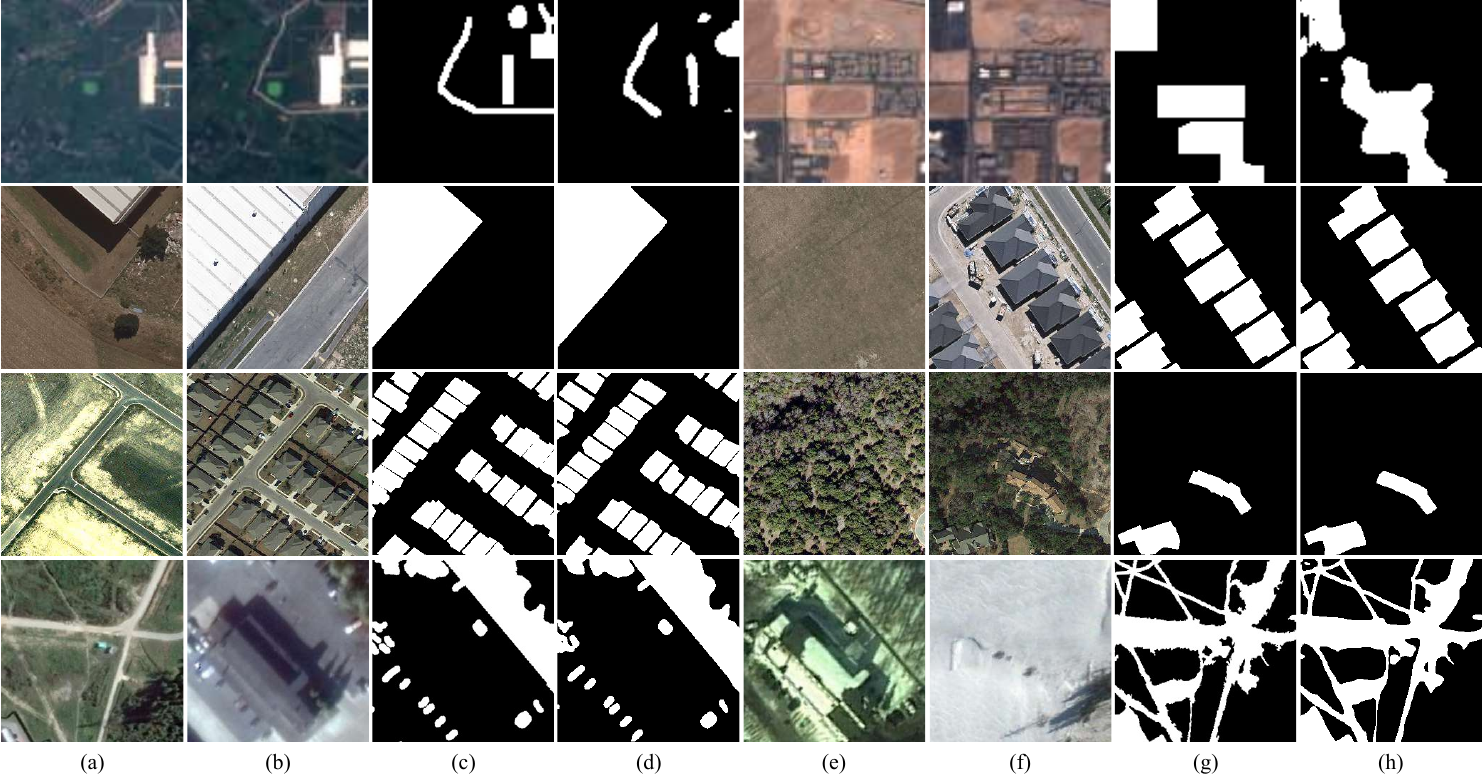}
  \caption{
  Visualization of the bi-temporal change detection predictions of MAE + MTP pretrained ViT-L + RVSA. The samples in four rows are from the testing sets of OSCD, WHU, LEVIR and SVCD/CDD, respectively. (a)(b)(e)(f) depict bi-temporal images of different samples, with (c) and (g) representing corresponding ground truth labels. Our prediction maps are shown at (d) and (h).
  }
  \label{bcd_predicts}
\end{figure*}

To further show the efficacy of MTP in enhancing RS foundation models, we present the predictions of MAE + MTP pretrained ViT-L + RVSA across detection, segmentation, and change detection tasks in Figure \ref{hdet_predicts}-\ref{bcd_predicts}. For detection, we demonstrate results across diverse scenes using horizontal or rotated bounding boxes. For segmentation, we display the original images alongside segmentation maps, highlighting building extraction masks in red. For change detection, we provide the bi-temporal images, ground truths, and predicted change maps. Our model accurately detects RS objects, extracts buildings, classifies land cover categories, and characterizes changes across diverse types. In summary, MTP enables the construction of an RS foundation model with over 300 parameters, which achieves superior representation capability for various downstream tasks.

\section{Conclusion}
\label{sec:conclusion}

In this paper, we introduce the multi-task pretraining (MTP) approach for building RS foundation models. MTP utilizes a shared encoder and task-specific decoder architecture to effectively pretrain convolutional neural networks and vision transformer backbones on three tasks: semantic segmentation, instance segmentation, and rotated object detection in a unified supervised learning framework. We evaluate MTP by examining the finetuning accuracy of these pretrained models on 14 datasets covering various downstream RS tasks. Our results demonstrate the competitive performance of these models compared to existing methods, even with larger models. Further experiments indicate that MTP excels in low-data finetuning scenarios but may offer diminishing returns with prolonged finetuning on large-scale datasets. We hope this research encourages further exploration of RS foundation models, especially in resource-constrained settings. Additionally, we anticipate the widespread application of these models across diverse fields of RS image interpretation due to their strong representation capabilities.

\section*{acknowledgement}

The numerical calculations in this paper are partly supported by the Dawning Information Industry Co., Ltd.

\bibliographystyle{ieeetr}
\bibliography{mtp}

\appendix

We present detailed finetuning accuracies of the three models, \textit{i.e.}, ViT-B + RVSA, ViT-L + RVSA, and InternImage-XL, on the DIOR, DIOR-R, FAIR1M-2.0, DOTA-V1.0, DOTA-V2.0, and LoveDA datasets in Table \ref{class_acc_dior}-\ref{class_acc_loveda}.

\begin{table*}[h]
  \caption{Detailed accuracies of different models on DIOR dataset.}
  \newcommand{\tabincell}[2]{\begin{tabular}{@{}#1@{}}#2\end{tabular}}
  \centering
  \resizebox{\linewidth}{!}{
  \begin{tabular}{lccccccc}
  \hline
    \bfseries Category &\bfseries  \tabincell{c}{ViT-B + RVSA \\w/o MTP} &\bfseries \tabincell{c}{ViT-B + RVSA\\ w MTP} & \bfseries \tabincell{c}{ViT-L + RVSA \\ w/o MTP} &\bfseries \tabincell{c}{ViT-L + RVSA\\ w MTP} & \bfseries \tabincell{c}{InternImage-XL \\ w/o MTP} & \bfseries \tabincell{c}{InternImage-XL\\ w MTP} \\
  \hline
  airplane & 68.2   &  87.5 & 76.5 & 93.8  &  65.0 & 69.0  \\
  airport &  91.2  & 92.1  & 92.6 & 91.6  & 91.3  & 92.8  \\
  baseballfield & 79.9   & 87.3  & 83.2 &  87.7  & 75.6  &  81.3 \\
  basketballcourt &  88.0  & 89.4  & 90.7 &  92.1 & 89.3  &  90.1 \\
  bridge  &  53.6  & 58.0  & 58.8 & 64.6  &  59.3 & 59.1  \\
  chimney & 82.1   &  83.7 & 84.1  & 85.9  &  84.9 &  84.8 \\
  expressway-service-area &  90.6  & 92.9  &  92.3 &  94.3 & 92.8  &  93.9 \\
  expressway-toll-station &  76.2  & 80.5  & 79.5  &  84.5 & 84.4  &  83.6 \\
  dam &  78.2  & 82.0  & 79.3  & 81.4   & 80.2  &  82.0 \\
  golffield &  84.9  & 88.1  & 85.7  &  87.3  & 86.2  & 83.9  \\
  groundtrackfield &  83.9  & 85.6  & 85.3  &  86.7  & 85.9  & 87.4  \\
  harbor & 56.8   & 62.4  & 60.4  & 64.5   &  62.2 &  63.3 \\
  overpass &  67.4  &  69.8 & 70.6  & 72.0  &  68.8  & 68.6  \\
  ship & 74.4  &  75.4 & 75.4  &  76.3 & 73.6  &  73.6 \\
  stadium &  82.8  &  85.8 &  84.3 & 85.6  & 83.7  &  85.5 \\
  storagetank & 61.4   & 62.5  & 65.3  & 62.6  &  59.6 & 57.7 \\
  tenniscourt  &  89.4  & 91.2  & 91.3 & 92.5  &  87.6 & 90.3  \\
  trainstation & 76.1   &  80.3 & 76.6 & 79.7  & 77.7  & 77.7 \\
  vehicle &  45.2  & 47.4  & 48.2 & 47.6  & 45.0  & 46.2 \\
  windmill &  85.0  & 86.5  &  85.1  & 90.2  &  89.3  &  89.3 \\
  \hline
  \bfseries mAP & 75.8  &  79.4 & 78.3 & 81.1  & 77.1  & 78.0  \\
  \hline
\end{tabular}
}
\label{class_acc_dior}
\end{table*}

\begin{table*}[h]
  \caption{Detailed accuracies of different models on DIOR-R dataset.}
  \newcommand{\tabincell}[2]{\begin{tabular}{@{}#1@{}}#2\end{tabular}}
  \centering
  \resizebox{\linewidth}{!}{
  \begin{tabular}{lccccccc}
  \hline
    \bfseries Category &\bfseries  \tabincell{c}{ViT-B + RVSA \\w/o MTP} &\bfseries \tabincell{c}{ViT-B + RVSA\\ w MTP} & \bfseries \tabincell{c}{ViT-L + RVSA \\ w/o MTP} &\bfseries \tabincell{c}{ViT-L + RVSA\\ w MTP} & \bfseries \tabincell{c}{InternImage-XL \\ w/o MTP} & \bfseries \tabincell{c}{InternImage-XL\\ w MTP} \\
  \hline
  airplane & 72.1  & 89.6  & 81.2  & 90.7  &  72.0  & 72.3 \\
  airport & 51.1  &  52.6 & 51.9  & 63.4  &  61.5 & 63.6 \\
  baseballfield & 80.8  & 81.2  & 81.1  & 90.0  &  80.6 & 80.9 \\
  basketballcourt & 81.3  &  87.8 & 90.1  & 90.1  &  90.0 & 90.1 \\
  bridge  & 44.9  & 48.6  & 48.1  & 56.4  &  53.5 & 54.7 \\
  chimney &  72.7 & 77.2  & 78.2  & 81.5  &  81.5 & 81.5 \\
  expressway-service-area &  87.5 & 89.1  & 88.4  &  89.4 &  89.9 & 89.7 \\
  expressway-toll-station & 69.3  &  71.6 & 74.7  &  80.1 &  79.5 & 79.6 \\
  dam & 35.5  & 43.3  & 39.8  & 39.9  &  43.0 & 45.9 \\
  golffield & 78.4  &  79.0 & 79.4  & 79.3  &  80.0 & 79.3 \\
  groundtrackfield &  81.9 & 84.2  & 84.3  & 85.1  &  85.2 & 85.3 \\
  harbor & 43.3  &  51.3 &  46.4 & 56.0  &  54.8 &  55.2 \\
  overpass &  60.1 &  60.9 & 60.5  & 67.2  &  64.2 & 65.8 \\
  ship & 81.2  &  81.2 &  81.2  & 81.1 &  81.3 & 81.3 \\
  stadium &  81.6 & 83.7  &  83.4 & 78.9  & 78.2  & 79.2 \\
  storagetank & 70.5  & 71.2  &  71.4 & 71.4  &  62.7 & 62.8 \\
  tenniscourt  & 89.2  & 90.2  & 90.0  & 90.4  & 81.5  & 90.1 \\
  trainstation &  65.6 & 66.6  &  65.2 & 73.9  &  66.7 & 67.5 \\
  vehicle &  49.3 & 50.5  & 51.0  & 51.8  &  50.5 & 51.1 \\
  windmill & 65.1   & 66.0  & 64.6  & 74.3  &  66.1 & 67.2 \\
  \hline
  \bfseries mAP & 68.1 & 71.3 & 70.5 & 74.5  & 71.1  & 72.2 \\
  \hline
\end{tabular}
}
\label{class_acc_diorr}
\end{table*}

\begin{table*}[t]
  \caption{Detailed accuracies of different models on FAIR1M-2.0 dataset.}
  \newcommand{\tabincell}[2]{\begin{tabular}{@{}#1@{}}#2\end{tabular}}
  \centering
  \resizebox{\linewidth}{!}{
  \begin{tabular}{lccccccc}
  \hline
    \bfseries Category &\bfseries  \tabincell{c}{ViT-B + RVSA \\w/o MTP} &\bfseries \tabincell{c}{ViT-B + RVSA\\ w MTP} & \bfseries \tabincell{c}{ViT-L + RVSA \\ w/o MTP} &\bfseries \tabincell{c}{ViT-L + RVSA\\ w MTP} & \bfseries \tabincell{c}{InternImage-XL \\ w/o MTP} & \bfseries \tabincell{c}{InternImage-XL\\ w MTP} \\
  \hline
  Boeing737 & 47.17 & 44.78  & 48.91 & 43.21 & 41.36 & 47.48 \\
  Boeing747 & 93.31 & 93.73 & 95.00 & 94.95 & 95.76 & 95.67 \\
  Boeing777 & 41.48 & 45.01 & 36.86 & 36.20 & 50.87 & 44.57 \\
  Boeing787 & 60.01 & 57.03 & 64.70 & 61.42 & 63.66 & 66.78 \\
  C919 & 42.43 & 44.27  & 33.94  & 50.14 &51.95  & 54.52 \\
  A220 & 53.64 & 52.23 & 56.51 & 52.70 &54.10  & 58.17 \\
  A321 & 71.56 & 72.12 & 72.98 & 72.68 & 69.31 & 70.76 \\
  A330 & 61.97 & 64.22 & 60.78 & 64.01 & 68.53 & 64.66 \\
  A350 & 75.76 & 73.14 & 75.94  & 70.90  & 80.19 & 78.89 \\
  ARJ21 & 20.36 &20.93  &  19.66 & 23.14  & 22.67 & 22.78 \\
  Passenger Ship &  20.12 &  21.29 & 20.38  & 19.80 & 15.76  & 15.55 \\
  Motorboat & 71.17 & 71.73 & 72.72 & 73.07 & 64.86 & 66.28 \\
  Fishing Boat & 35.42 & 36.30 & 36.30 & 33.94 & 28.38 & 31.34 \\
  Tugboat & 31.12 & 32.66 & 36.00 & 32.52 &  27.33 &  25.38\\
  Engineering Ship & 22.75 & 25.90 & 29.02 & 28.63 &20.16  & 21.00 \\
  Liquid Cargo Ship &48.73  & 49.30 & 52.80  & 49.31 & 46.91 &  46.32 \\
  Dry Cargo Ship & 53.07 & 53.12 & 53.87  & 51.09 & 49.18 & 49.13 \\
  Warship & 38.17 & 40.88 & 45.66  & 43.44 & 33.45  & 38.04 \\
  Small Car & 76.77 &  76.98 & 77.65  & 77.23 & 72.77 &  72.92\\
  Bus & 45.60 & 42.16 & 51.73  & 51.73 & 47.59 & 46.79 \\
  Cargo Truck &59.64  & 59.87 & 61.56  &60.53 & 56.15 & 56.32 \\
  Dump Truck & 61.73 & 61.85 & 63.08  & 60.95 & 57.69 &57.48  \\
  Van & 77.08 & 77.33 &  78.22 & 77.74 & 73.00 & 73.08 \\
  Trailer &19.74  &19.34  &  23.63 & 22.48 & 14.48 & 17.55 \\
  Tractor &  1.33 &  1.79 & 2.21  & 1.83 & 0.71 & 1.23  \\
  Excavator & 23.38 &  25.03 &  27.58 & 29.12 & 21.49 & 18.68 \\
  Truck Tractor & 50.00 & 49.83  &  48.71 &  52.15&50.62  & 48.44 \\
  Basketball Court & 64.50 & 63.35  & 63.46  &64.40  & 60.69 & 60.46 \\
  Tennis Court & 91.45 & 90.71  &  92.09 & 91.53  & 89.07 & 90.22 \\
  Football Field & 66.23 & 67.21  & 70.72  & 70.75 & 68.44 &  68.98 \\
  Baseball Field &  91.81 & 91.52  & 92.27  & 91.80 & 87.93 & 88.78 \\
  Intersection &63.65  &  64.73 &  64.94 &  66.22 & 64.30 & 63.28 \\
  Roundabout & 26.13 &25.43 & 28.36 & 31.00 & 30.71 &  27.07 \\
  Bridge & 45.91 & 49.38 & 50.66 & 51.32 & 42.81 & 43.33  \\
  \hline
  \bfseries mAP &51.56 & 51.92 & 53.20  & 53.00 & 50.67  & 50.94 \\
  \hline
\end{tabular}
}
\label{class_acc_fair}
\end{table*}

\begin{table*}[t]
  \caption{Detailed accuracies of different models on DOTA-V1 dataset.}
  \newcommand{\tabincell}[2]{\begin{tabular}{@{}#1@{}}#2\end{tabular}}
  \centering
  \resizebox{\linewidth}{!}{
  \begin{tabular}{lccccccc}
  \hline
    \bfseries Category &\bfseries  \tabincell{c}{ViT-B + RVSA \\w/o MTP} &\bfseries \tabincell{c}{ViT-B + RVSA\\ w MTP} & \bfseries \tabincell{c}{ViT-L + RVSA \\ w/o MTP} &\bfseries \tabincell{c}{ViT-L + RVSA\\ w MTP} & \bfseries \tabincell{c}{InternImage-XL \\ w/o MTP} & \bfseries \tabincell{c}{InternImage-XL\\ w MTP} \\
  \hline
  plane & 88.42 & 88.91  & 88.52 & 88.33 & 88.91 & 88.96 \\
  baseball-diamond & 85.03 & 84.07  & 85.36 & 86.59 & 86.78 & 85.93 \\
  bridge & 60.86 & 60.76 & 61.55 & 63.38 & 59.93 & 60.62 \\
  ground-track-field & 82.39 & 82.93  & 81.25  & 83.49  & 81.05 &  81.26 \\
  small-vehicle &80.70  & 80.41 & 80.69 & 81.06 & 80.80 & 80.08 \\
  large-vehicle & 85.76 & 86.16 & 86.44 & 86.48 & 85.06 & 84.55 \\
  ship & 88.58 & 88.64 & 88.51  & 88.53 & 88.38 & 87.96 \\
  tennis-court &90.88  & 90.87 & 90.87 & 90.87  & 90.81  &90.86  \\
  basketball-court & 86.61 & 86.21 & 85.80  & 86.26   & 86.27  & 86.37 \\
  storage-tank & 86.88  & 86.76  & 86.84 & 85.80  & 86.19 & 86.50 \\
  soccer-ball-field & 63.79 &  63.91 & 69.81  & 67.17  & 69.64  & 68.93 \\
  roundabout & 72.52 & 71.00  & 72.51 & 71.84 & 71.50 & 73.11 \\
  harbor & 78.52 & 79.04 & 84.82  & 84.94 & 79.21 & 78.76 \\
  swimming-pool & 80.53 & 82.17 & 79.99 & 81.93 & 81.50 & 80.98 \\
  helicopter & 81.00 & 78.34 & 78.51 & 78.31 & 67.57 & 76.63 \\
  \hline
  \bfseries mAP & 80.83 & 80.68 & 81.43 &  81.66 &  80.24 & 80.77  \\
  \hline
\end{tabular}
}
\label{class_acc_dota1}
\end{table*}

\begin{table*}[t]
  \caption{Detailed accuracies of different models on DOTA-V2 dataset.}
  \newcommand{\tabincell}[2]{\begin{tabular}{@{}#1@{}}#2\end{tabular}}
  \centering
  \resizebox{\linewidth}{!}{
  \begin{tabular}{lccccccc}
  \hline
    \bfseries Category &\bfseries  \tabincell{c}{ViT-B + RVSA \\w/o MTP} &\bfseries \tabincell{c}{ViT-B + RVSA\\ w MTP} & \bfseries \tabincell{c}{ViT-L + RVSA \\ w/o MTP} &\bfseries \tabincell{c}{ViT-L + RVSA\\ w MTP} & \bfseries \tabincell{c}{InternImage-XL \\ w/o MTP} & \bfseries \tabincell{c}{InternImage-XL\\ w MTP} \\
  \hline
  plane & 77.86 & 78.30 & 79.16  & 78.57 & 78.52 & 70.98 \\
  baseball-diamond & 48.99 & 52.58 & 48.16  & 45.54 & 50.96 & 50.82 \\
  bridge & 46.55 & 48.42 & 47.97 & 49.72 & 43.21 & 43.60 \\
  ground-track-field & 63.10 & 59.05 & 61.57 & 56.42 & 59.77 &  59.25\\
  small-vehicle & 43.55  & 43.65 & 43.74  & 43.78 & 43.58 & 43.55 \\
  large-vehicle & 56.85 & 57.15  & 61.14  & 62.26  & 56.11  & 56.50 \\
  ship & 61.09 & 61.08 & 68.60  & 68.76 & 61.38 &  61.41 \\
  tennis-court & 76.90 & 77.83  & 78.45 & 74.89 & 77.61 & 78.23 \\
  basketball-court & 54.57 & 56.32 & 61.97  & 64.17  & 58.20 & 61.41 \\
  storage-tank & 58.55 & 59.23  & 59.62 & 58.70 & 58.55 & 51.30 \\
  soccer-ball-field & 36.37 & 36.93 & 45.98 & 43.70 & 48.88 & 42.89 \\
  roundabout & 50.39 & 51.26 & 54.89 & 49.46 & 50.17  & 50.60 \\
  harbor & 56.34 & 56.97 &62.03  & 63.32 & 57.86 & 56.84 \\
  swimming-pool & 63.89 & 63.05 & 64.34 & 64.59 & 58.43 & 58.31 \\
  helicopter & 65.43  & 66.40 & 70.23 & 72.71  & 58.16  & 59.55 \\
  container-crane & 39.19 & 44.24 & 46.94 & 50.91 & 40.58  & 49.21  \\
  airport & 79.90 & 87.57 & 87.64 & 87.64 & 77.39 & 84.73 \\
  helipad & 14.43  & 9.33 & 18.92 & 16.27 & 7.91 & 13.09 \\
  \hline
  \bfseries mAP & 55.22 & 56.08 & 58.96  & 58.41  & 54.85  & 55.13 \\
  \hline
\end{tabular}
}
\label{class_acc_dota2}
\end{table*}

\begin{table*}[h]
  \caption{Detailed accuracies of different models on LoveDA dataset.}
  \newcommand{\tabincell}[2]{\begin{tabular}{@{}#1@{}}#2\end{tabular}}
  \centering
  \resizebox{\linewidth}{!}{
  \begin{tabular}{lccccccc}
  \hline
    \bfseries Category &\bfseries  \tabincell{c}{ViT-B + RVSA \\w/o MTP} &\bfseries \tabincell{c}{ViT-B + RVSA\\ w MTP} & \bfseries \tabincell{c}{ViT-L + RVSA \\ w/o MTP} &\bfseries \tabincell{c}{ViT-L + RVSA\\ w MTP} & \bfseries \tabincell{c}{InternImage-XL \\ w/o MTP} & \bfseries \tabincell{c}{InternImage-XL\\ w MTP} \\
  \hline
  background & 45.91 & 45.92 & 47.26 & 47.14 & 46.63 & 46.80 \\
  building & 57.93  & 59.40 & 59.27 & 62.69 & 61.98 & 62.60 \\
  Road & 56.08 & 56.15 & 59.54 &58.00  & 58.25  & 58.96  \\
  water & 79.72 & 80.66 & 81.45 & 81.43 & 82.14 &  82.25\\
  barren & 16.49 &16.56  & 17.59 & 19.27 & 18.11 & 17.49 \\
  forest & 46.03 & 46.38 & 47.39 & 46.82  & 47.99  & 47.63 \\
  agriculture & 61.48 & 61.67 & 63.55 & 63.80 & 62.40  & 63.44  \\
  \hline
  \bfseries mIOU & 51.95 & 52.39 & 53.72 & 54.17 & 53.93 & 54.17 \\
  \hline
\end{tabular}
}
\label{class_acc_loveda}
\end{table*}

\end{document}